\PassOptionsToPackage{table}{xcolor}
\documentclass[11pt]{article}

\usepackage[preprint]{acl}

\usepackage{times}
\usepackage{latexsym}
\usepackage[T1]{fontenc}
\usepackage[utf8]{inputenc}
\usepackage{microtype}
\usepackage{inconsolata}
\usepackage{natbib}

\usepackage{adjustbox}
\usepackage{soul}
\usepackage{url}
\usepackage{graphicx}
\usepackage{amsmath}

\usepackage{amssymb}
\usepackage{amsthm}
\usepackage{booktabs}
\usepackage{algorithm}
\usepackage{algorithmic}
\usepackage{longtable}
\usepackage{multirow}
\usepackage{multicol}
\usepackage{makecell}
\usepackage[table]{xcolor}
\usepackage{float}
\usepackage{tabularx}
\usepackage{subcaption}
\usepackage{placeins}
\usepackage{xspace}

\usepackage[most]{tcolorbox}
\newtcolorbox{promptbox}[2][]{
    colback=blue!5!white,
    colframe=blue!50!black,
    colbacktitle=blue!50!black,
    coltitle=white,
    fonttitle=\bfseries\large,
    title={#2},
    sharp corners=south,
    boxrule=0.5pt,
    left=5pt, right=5pt, top=5pt, bottom=5pt,
    #1
}
\usepackage{pifont}

\usepackage{enumitem}
\setlist{nosep}

\definecolor{darkgreen}{RGB}{30, 150, 30}
\definecolor{darkheader}{RGB}{64, 64, 64}
\definecolor{lightbody}{RGB}{242, 242, 242}
\definecolor{slbeige}{RGB}{245, 240, 230}
\definecolor{slblue}{RGB}{225, 235, 245}
\definecolor{slorange}{RGB}{250, 235, 220}
\definecolor{softred}{RGB}{200, 50, 50}
\newcommand{\cmark}{\textcolor{darkgreen}{\ding{51}}}
\newcommand{\xmark}{\textcolor{softred}{\ding{55}}}

\newcommand{\icon}[1]{%
\adjustbox{valign=c}{\includegraphics[height=1.2em]{#1}}%
}

\newcommand{\raaf}{RA-AlignF1\xspace}
\newcommand{\tmax}{\ensuremath{\tau}-MaxSim\xspace}
\newcommand{\raafm}{\text{RA-AlignF1}}
\newcommand{\tmaxm}{\tau\text{-MaxSim}}

\title{SurveyLens: A Discipline-Aware Benchmark for Automatic \\ Survey Generation}

\author{
  \textbf{Beichen Guo\textsuperscript{1}\thanks{\ Both authors contributed equally to this research.}},
  \textbf{Zhiyuan Wen\textsuperscript{1}\footnotemark[1]},
  \textbf{Jia Gu\textsuperscript{1}},
  \textbf{Haochen Shi\textsuperscript{1}},
  \textbf{Jian Wang\textsuperscript{2}},
  \\
  \textbf{Senzhang Wang\textsuperscript{3}},
  \textbf{Haoyang Li\textsuperscript{1}},
  \textbf{Ruosong Yang\textsuperscript{1}},
  \textbf{Shuaiqi Liu\textsuperscript{4}}
\\
  \textsuperscript{1}The Hong Kong Polytechnic University \\
  \textsuperscript{2}Sichuan University \\
  \textsuperscript{3}Central South University \\
  \textsuperscript{4}Alibaba Cloud \\
  \small \texttt{beichen.guo@connect.polyu.hk}, \texttt{zhiyuan.wen@polyu.edu.hk}
}

\begin{document}
\maketitle

\begin{abstract}
Automatic Survey Generation (ASG) aims to produce comprehensive literature surveys by retrieving, organizing, and synthesizing academic papers. Despite rapid progress in specialized ASG frameworks and Deep Research agents, existing evaluations largely center on Computer Science or rely on generic criteria, leaving it unclear whether current systems satisfy the survey standards of diverse disciplines. We introduce \textbf{SurveyLens}, the first discipline-aware ASG benchmark. SurveyLens comprises SurveyLens-1k, a curated dataset of 1,000 human-written surveys across 10 disciplines, and a dual-lens framework that combines discipline-aware rubric scoring with reference-based alignment to human-written surveys. Evaluating 11 state-of-the-art systems across vanilla LLMs, ASG systems, and Deep Research agents (Figure~\ref{fig:radar_intro}), we find that Deep Research agents are the only paradigm robust across all 10 disciplines, ASG systems lead on structural planning, and all paradigms remain weak on reference quality, providing practical guidance for discipline-specific tool selection and future ASG design.\footnote{Dataset, code, results, and the annotation platform are released at \url{https://github.com/TechnicolorGUO/SurveyLens}.}
\end{abstract}

\section{Introduction}
\begin{table*}[th!]
    \centering
    \small
    \resizebox{\textwidth}{!}{%
    \begin{tabular}{lccccccccccc}
        \toprule
        \multirow{3}{*}{\textbf{Assessment Methods}} & 
        \multirow{3}{*}{\textbf{Disciplines}} & 
        \multirow{3}{*}{\textbf{\#Surveys}} & 
        \multirow{3}{*}{\textbf{\# Systems}} &
        \multicolumn{6}{c}{\textbf{Assessment Components}} & 
        \multirow{3}{*}{\textbf{\shortstack{Human\\Eval.}}} & 
        \multirow{3}{*}{\textbf{\shortstack{Discipline-Aware\\Assessment}}} \\
        
        \cmidrule(lr){5-10}
        
         & & & & 
        \multicolumn{2}{c}{\textbf{Outline}} & 
        \multicolumn{2}{c}{\textbf{Content}} & 
        \multicolumn{2}{c}{\textbf{Reference}} & 
         & \\
        
        \cmidrule(lr){5-6} \cmidrule(lr){7-8} \cmidrule(lr){9-10}
        
         & & & & 
        \textbf{Ref.} & \textbf{Free} & 
        \textbf{Ref.} & \textbf{Free} & 
        \textbf{Ref.} & \textbf{Free} & 
         & \\
        \midrule
        
        
        SciReviewGen~\cite{kasanishi_scireviewgen_2023} & CS Only & 10K & 
        6 & 
        \xmark & \xmark & 
        \cmark & \xmark & 
        \xmark & \xmark & 
        \cmark & 
        \xmark \\
        
        LLM$\times$MapReduce-V2~\cite{wang_llmtimesmapreduce-v2_2025} & CS Only & 384 & 
        4 & 
        \xmark & \xmark & 
        \xmark & \cmark & 
        \cmark & \xmark & 
        \cmark & 
        \xmark \\

        SurveyForge~\cite{yan_surveyforge_2025} & CS Only & 100 & 
        3 & 
        \xmark & \cmark & 
        \xmark & \cmark & 
        \cmark & \xmark & 
        \cmark & 
        \xmark \\

        SciSage~\cite{shi_scisage_2025} & CS Only & 46 & 
        4 & 
        \xmark & \cmark & 
        \xmark & \cmark & 
        \cmark & \xmark & 
        \cmark & 
        \xmark \\
        
        SurveyGen~\cite{bao_surveygen_2025} & 4 Discip. & 4.2K & 
        3 & 
        \cmark & \cmark & 
        \cmark & \xmark & 
        \cmark & \xmark & 
        \cmark & 
        \xmark \\
        
        SurveyBench~\cite{sun_surveybench_2025} & CS Only & 4.9K & 
        4 & 
        \xmark & \cmark & 
        \cmark & \cmark & 
        \xmark & \xmark & 
        \xmark & 
        \xmark \\

        SurveyEval~\cite{zhao_surveyeval_2025} & 7 Discip. & N/A & 
        7 & 
        \xmark & \cmark & 
        \xmark & \cmark & 
        \cmark & \xmark & 
        \xmark & 
        \xmark \\

        SurGE~\cite{su_surge_2026} & CS Only & 205 & 
        3 & 
        \cmark & \cmark & 
        \cmark & \cmark & 
        \xmark & \cmark & 
        \xmark & 
        \xmark \\


        \midrule
        
        \rowcolor{gray!10} 
        \textbf{SurveyLens (Ours)} & 
        \textbf{10 Discip.} & 
        \textbf{1,000} & 
        \textbf{11} & 
        \cmark & \cmark & 
        \cmark & \cmark & 
        \cmark & \cmark & 
        \cmark & 
        \cmark \\
        \bottomrule
    \end{tabular}
    }
    \caption{Comparison of \textbf{SurveyLens} with existing ASG assessment methods.}
    \label{tab:comparison}
    \vspace{-6pt}
\end{table*}
The rapid growth of scientific literature has made it increasingly difficult for researchers to keep up with new developments through manual literature review. Automatic Survey Generation (ASG) addresses this challenge by combining Large Language Models (LLMs) with Retrieval-Augmented Generation (RAG)~\cite{lewis_retrieval-augmented_2020} to retrieve relevant papers and synthesize them into survey-style narratives. Existing ASG systems have evolved from pipeline-based approaches that streamline discrete literature review tasks~\cite{wang_autosurvey_2024,liang_surveyx_2025,wang_llmtimesmapreduce-v2_2025}  to human-in-the-loop paradigms enabling iterative refinement~\cite{wen_interactivesurvey_2025,chao_llmmapreduce-v3_2025,zhang_deep_2025}, and finally to multi-agent systems that simulate collaborative survey workflows~\cite{liu_agentic_2025,shi_scisage_2025}. In parallel, commercial Deep Research products such as Gemini Deep Research~\cite{gemini_gemini_2025} and Qwen Deep Research~\cite{qwen_qwen_2025} have emerged as general-purpose alternatives.

\begin{figure}[t!]
    \centering
    \includegraphics[width=0.9\linewidth]{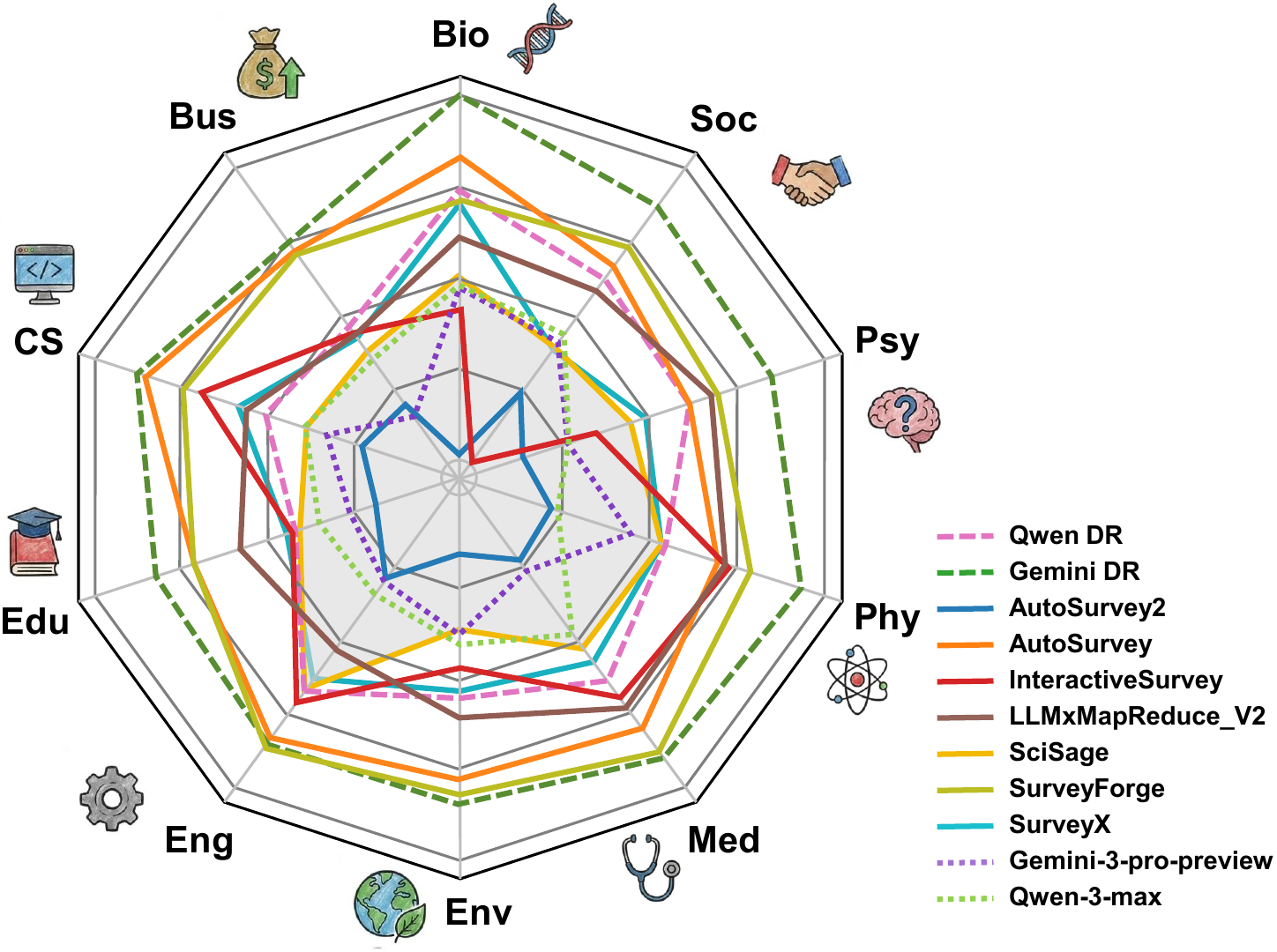}
    \caption{Performance comparison of vanilla LLMs, ASG systems, and deep research agents in generating surveys across 10 research disciplines.}
    \label{fig:radar_intro}
    \vspace{-8pt}
\end{figure}

Despite this progress, ASG evaluation remains inadequate along two axes: \emph{what} surveys are evaluated on and \emph{how} they are evaluated. On the data side, existing benchmarks~\cite{kasanishi_scireviewgen_2023,wang_llmtimesmapreduce-v2_2025,yan_surveyforge_2025,shi_scisage_2025,sun_surveybench_2025,su_surge_2026} are predominantly restricted to Computer Science (CS) and overlook how writing conventions diverge across fields: a physics survey is anchored in equations and derivations, while a sociology review proceeds through qualitative narrative synthesis. General discipline-aware LLM benchmarks such as SuperGPQA~\cite{team_supergpqa_2025} do not close this gap either, as they probe atomic knowledge through Question Answering (QA) rather than the planning and coherent narrative construction required for survey writing. On the methodology side, semantic metrics~\cite{kasanishi_scireviewgen_2023} such as ROUGE~\cite{lin_rouge_2004} and BERTScore~\cite{zhang_bertscore_2020} fail to capture the structural diversity of survey writing, while recent approaches~\cite{wang_autosurvey_2024,wang_llmtimesmapreduce-v2_2025} based on LLM-as-a-judge~\cite{zheng_judging_2023} rely on generic criteria that lack interpretability and disciplinary grounding.



To bridge these gaps, we introduce \textbf{SurveyLens}, a discipline-aware benchmark for ASG\@. On the data side, we construct \textit{SurveyLens-1k}, a dataset of 1,000 human-written surveys evenly spanning 10 academic disciplines. On the methodology side, we design a dual-lens framework: (i) \textit{Discipline-Aware Rubric Evaluation} (reference-free), which uses an LLM-as-a-judge with field-specific rubrics distilled bottom-up from SurveyLens-1k and aspect weights calibrated to expert preferences via Bradley-Terry modeling; and (ii) \textit{Human-Reference Alignment Evaluation} (reference-based), which compares against paired human-written surveys through two metrics, Redundancy-Aware Alignment F1 (strict one-to-one Hungarian matching with a redundancy penalty, targeting paragraph bloat) and Thresholded MaxSim (margin-based max-similarity without the matching constraint, quantifying the degree of alignment).

Our analysis reveals that no single paradigm dominates: Deep Research agents
are the only paradigm robust across all 10 disciplines, ASG systems and DR
agents split along a structure-vs-content axis, and reference quality remains
a universal bottleneck. We summarize our contributions as follows:
\begin{itemize}
    \item \textbf{High-Quality Survey Dataset.}
    We construct SurveyLens-1k, a curated dataset of 1,000
    human-written surveys evenly distributed across 10 academic disciplines,
    with each survey decomposed into outline, content, and references for
    component-level evaluation.

    \item \textbf{Discipline-Aware ASG Evaluation.}
    We propose a dual-lens framework that combines reference-free
    discipline-aware rubrics with human-reference alignment metrics to assess
    both disciplinary writing norms and coverage of human-written surveys, and
    validate it against judgments from 30 PhD-level domain experts across
    all 10 disciplines.

    \item \textbf{Cross-Paradigm Empirical Analysis.}
    We benchmark 11 ASG methods spanning vanilla LLMs, specialized ASG
    systems, and Deep Research Agents, and translate the resulting
    paradigm-level patterns into actionable guidance for discipline-aware
    system selection.
\end{itemize}

\section{Related Work}
Recent ASG systems span pipeline-based approaches~\cite{wang_autosurvey_2024,liang_surveyx_2025,yan_surveyforge_2025,wang_llmtimesmapreduce-v2_2025,wu_autosurvey2_2025}, human-in-the-loop paradigms~\cite{wen_interactivesurvey_2025,chao_llmmapreduce-v3_2025,zhang_deep_2025}, multi-agent frameworks~\cite{shi_scisage_2025,liu_agentic_2025,nguye_surveyg_2025}, and commercial Deep Research agents~\cite{openai_introducing_2025,gemini_gemini_2025,qwen_qwen_2025}, yet their training corpora, retrieval pools, and self-reported benchmarks are overwhelmingly CS-centered, leaving open whether architectural conclusions transfer across disciplines.

Prior ASG assessment spans n-gram overlap~\cite{kasanishi_scireviewgen_2023}, LLM-as-a-judge content frameworks~\cite{wang_autosurvey_2024,liang_surveyx_2025}, and structural metrics~\cite{zhu_hierarchical_2023,lai_instruct_2024,bao_surveygen_2025,su_surge_2026}, with recent benchmarks~\cite{wang_llmtimesmapreduce-v2_2025,yan_surveyforge_2025,su_surge_2026,sun_surveybench_2025,bao_surveygen_2025,zhao_surveyeval_2025,guo_sgsimeval_2026} expanding scale and coverage. Two gaps persist: (i) they rely on discipline-agnostic criteria that overlook field-specific survey-writing conventions; (ii) human validation, when reported, draws exclusively on CS experts, leaving non-CS rubrics empirically unverified. As summarized in Table~\ref{tab:comparison}, prior benchmarks typically expand along a single axis such as scale, system coverage, or metric type, while leaving discipline-aware assessment incomplete across survey components and evaluation settings. SurveyLens is the only benchmark covering all six (Outline/Content/Reference)$\times$(reference-free/reference-based) cells, while also spanning 10 disciplines, 11 ASG baselines, and discipline-aware human evaluation.


\section{SurveyLens}
In this section, we introduce SurveyLens, covering (a) the construction of SurveyLens-1k, (b) Structured Survey Representation, and (c) the dual-lens evaluation framework.

\subsection{SurveyLens-1k Construction}

\begin{figure*}[htbp]
    \centering
    \includegraphics[width=\textwidth]{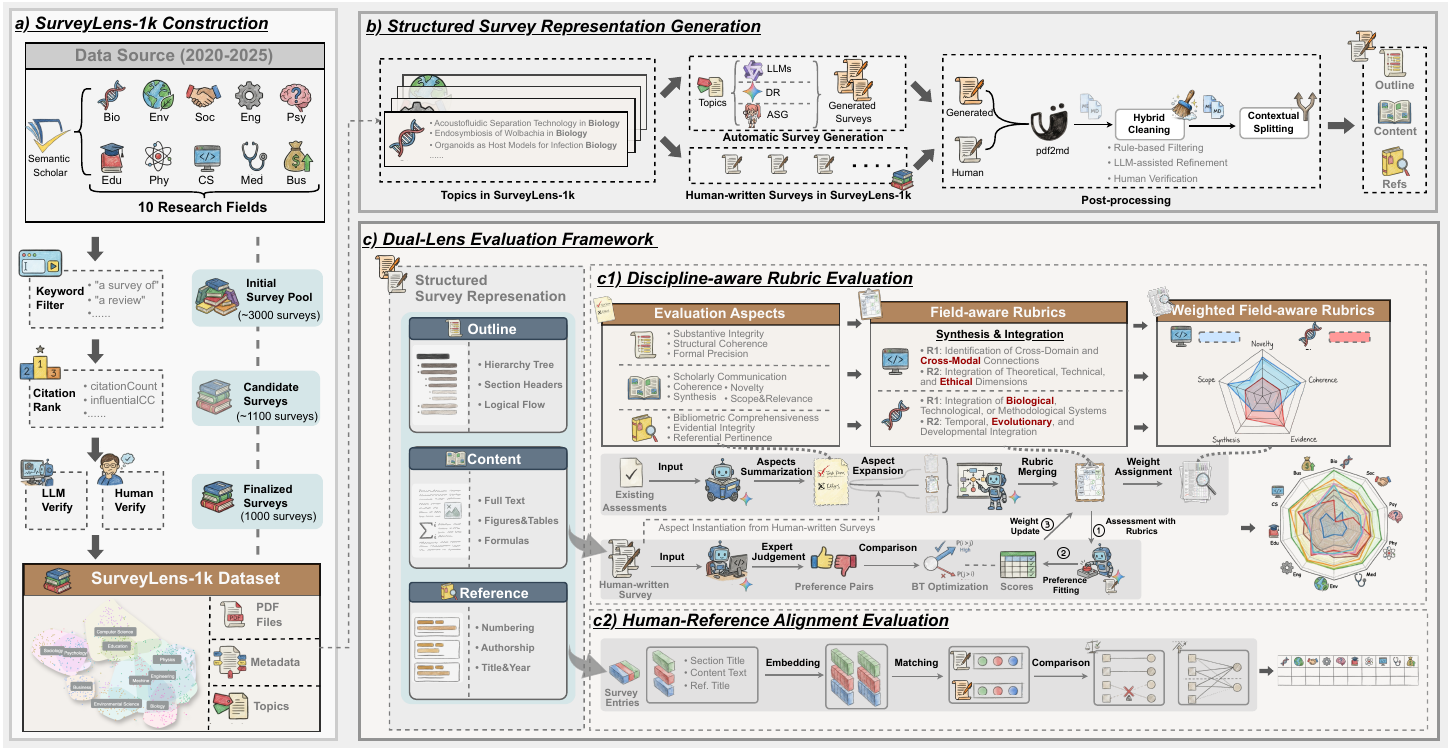}
    \caption{Overview of the proposed SurveyLens framework.}
    \label{fig:pipeline}
    \vspace{-6pt}
\end{figure*}

To ensure robust field-aware assessment of ASG methods, we constructed \textbf{SurveyLens-1k}, a large-scale dataset capturing the structural and content variances inherent to different fields, comprising 1,000 high-quality, human-written survey papers across 10 academic disciplines, with 100 surveys per field, as shown in Figure~\ref{fig:pipeline}. Papers are sourced from Semantic Scholar via dual citation-based ranking (raw and influence-weighted) and validated through hybrid LLM-plus-human filtering; full construction details are in Appendix~\ref{app:dataset_construction}.

\begin{figure}[htbp]
\centering
\includegraphics[width=\columnwidth]{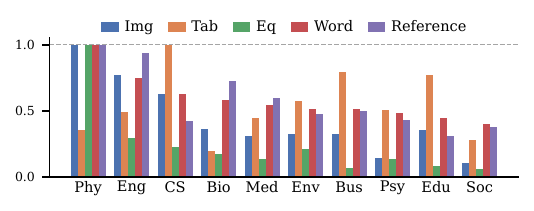}
\caption{Cross-disciplinary variance in SurveyLens-1k: per-paper averages of figures (Img), tables (Tab), equations (Eq), word count (Word), and references, each normalized to its per-metric maximum.}
\label{fig:dataset_variance}
\vspace{-6pt}
\end{figure}

Per-paper structural statistics reveal substantial cross-disciplinary variation (Figure~\ref{fig:dataset_variance}). Physics is the most technically dense field, leading on equations (\(411.4\)/paper), word count (\(23.4\)k), figures, and references. Tables are the only axis where CS overtakes Physics (\(8.7\) vs.\ \(3.1\)/paper), reflecting CS's reliance on tabular comparisons. Variance reaches up to an order of magnitude across the corpus (e.g., \(\sim 16\times\) on equations); the full per-discipline breakdown is given in Table~\ref{tab:dataset_stats} (Appendix~\ref{app:dataset_stats}). These systematic differences motivate our field-aware evaluation framework as well.
\subsection{Structured Survey Representation}

To enable component-wise evaluation (\S\ref{sec:dual_lens}), we adopt a unified \textit{Structured Survey Representation} for both human-written references in SurveyLens-1k and ASG system outputs, formally defined as the triplet
\begin{equation}
\mathcal{S} = (\mathcal{O}, \mathcal{C}, \mathcal{R}),
\end{equation}
where $\mathcal{O}$ denotes the hierarchical \textbf{outline} (i.e., nested section titles), $\mathcal{C}$ represents the \textbf{content} (including textual paragraphs, figures, tables, formulas, and in-text citations), and $\mathcal{R}$ is the \textbf{reference list} (containing bibliographic metadata such as titles and publication venues). We obtain $\mathcal{S}$ from raw PDFs via a four-stage pipeline: MinerU~\cite{wang_mineru_2024} PDF parsing, rule-based hierarchy and reference extraction, LLM-based normalization, and human verification (detailed in Appendix~\ref{app:ssr_pipeline}).

\subsection{Dual-Lens Evaluation Framework}\label{sec:dual_lens}

Each component of $\mathcal{S}$ is scored by a reference-free rubric lens and a reference-based alignment lens.

\subsubsection{Discipline-Aware Rubric Evaluation}
\label{sec:rubric}
Existing assessment methods apply universal criteria and overlook field-specific writing conventions. We therefore distill universal evaluation aspects from existing methods (Appendix~\ref{app:aspect_catalog}, Table~\ref{tab:aggregated_aspects}; aggregation prompt in Figure~\ref{fig:prompt_aggregate}), instantiate them into \textit{discipline-aware rubrics} using SurveyLens-1k, and learn \textit{preference-aligned weights} from expert judgments to calibrate the relative importance of indicators.

\paragraph{Discipline-Aware Rubric Generation.}
For each component $c \in \{\mathcal{O}, \mathcal{C}, \mathcal{R}\}$, we define a set of aspects $\mathcal{A}_c$ (synthesized from existing assessments) and construct rubrics via a two-stage process:

\begin{equation}
a \in \mathcal{A}_c
\;\xrightarrow[\text{per survey}]{\text{expand}}
\mathcal{K}^{\prime}_{d,c,a}
\;\xrightarrow[\text{per discipline}]{\text{merge}}
\mathcal{K}_{d,c,a}.
\end{equation}
The expand stage instantiates each aspect for human-written surveys in discipline $d$, while the merge stage consolidates them into shared rubric items $\mathcal{K}_{d,c,a}$. Prompts are shown in Figures~\ref{fig:prompt_expand} and~\ref{fig:prompt_merge}.

\paragraph{Expert-Preference-Aligned Weighting.}
We adopt the Bradley-Terry (BT) model to convert pairwise preferences into per-aspect weights, with preferences obtained from an LLM proxy (\textit{Gemini-3-Pro}, double round-robin with position swapping; validated against genuine expert judgments in \S\ref{sec:human_eval})~\cite{zheng_judging_2023,wang_large_2024}. Given normalized aspect scores $\bar{x}_{i,k} = x_{i,k}/x_{\max} \in [0.2,1]$ ($x_{\max}=5$, judge scores satisfy $x_{i,k}\geq 1$) and pairwise preferences $\mathcal{P} = \{(s_i, s_j): s_i \succ s_j\}$, we estimate weights $\mathbf{w}$ by maximizing:
\begin{equation}
\small
\hat{\mathbf{w}} = \arg\max_{\mathbf{w} \geq \mathbf{0}} \sum_{\mathcal{P}} \log \sigma\bigl(\mathbf{w}^\top (\bar{\mathbf{x}}_i - \bar{\mathbf{x}}_j)\bigr) - \alpha \|\mathbf{w}\|_2^2
\end{equation}

We solve Eq.~(3) per $(d, c)$ pair and normalize to $\tilde{w}_{c,k} = \hat{w}_{c,k}/\sum_{k'} \hat{w}_{c,k'}$. The final compliance score is $\mathcal{S}^{\textsc{rub}}_{d,c}(s_i) = 5\sum_k \tilde{w}_{c,k}\,\bar{x}_{i,k} \in [1,5]$, expressed on a 1--5 scale (Appendix~\ref{sec:scoring_example}).
The resulting discipline-aware rubric is validated against judgments from $30$ PhD-level domain experts spanning all $10$ disciplines (\S\ref{sec:human_eval}; Appendix~\ref{app:annotation_platform},~\ref{app:full_human_alignment}).

\subsubsection{Human-Reference Alignment Evaluation}

We use each paired human-written survey as one valid reference among many for the same topic; this lens complements rubric scoring by measuring local semantic correspondence. Conventional reference-based metrics such as ROUGE and embedding-based F1 suffer from two limitations: (1) they rely on many-to-many matching that ignores redundancy in generated content, and (2) they apply uniform thresholds that do not account for component-specific similarity distributions~\cite{bao_surveygen_2025,zhao_surveyeval_2025}. To overcome these issues, we propose two entry-level semantic alignment scores adapted for long-form survey evaluation: \raaf (Redundancy-Aware Alignment F1) and \tmax (Thresholded MaxSim). \raaf follows the precision--recall--F1 tradition of semantic matching metrics such as BERTScore~\cite{zhang_bertscore_2020}, additionally enforcing one-to-one Hungarian assignment and applying a redundancy penalty to target the paragraph-bloat failure mode of ASG systems. \tmax adopts a MaxSim-style operator~\cite{khattab_colbert_2020} at the entry level, introducing a component-specific threshold margin that quantifies the \emph{degree} of alignment rather than a binary threshold decision.

Instead of taking the full component as the input, we split the components into entries for fine-grained comparison. For each component-topic pair $(c,t)$, we define the generated entry set $\mathcal{E}_{c,t}$ and ground-truth entry set $\mathcal{G}_{c,t}$ as:
\begin{equation}
\begin{aligned}
\mathcal{E}_{c,t} &= \{ e_{c,t}^{(1)}, e_{c,t}^{(2)}, \dots, e_{c,t}^{(m_{c,t})} \}, \\
\mathcal{G}_{c,t} &= \{ g_{c,t}^{(1)}, g_{c,t}^{(2)}, \dots, g_{c,t}^{(n_{c,t})} \},
\end{aligned}
\end{equation}
where $m_{c,t}$ and $n_{c,t}$ denote the number of entries for the generated and ground-truth human-written surveys in SurveyLens-1k, respectively. Concretely, the similarity-bearing content of each entry is the section title for $\mathcal{O}$, the section body for $\mathcal{C}$, and the reference title for $\mathcal{R}$; structural metadata (hierarchical level, heading, additional bibliographic fields) is retained alongside each entry but does not enter the similarity computation.

\paragraph{Redundancy-Aware Alignment F1 (\raaf).}
\raaf measures overall coverage and faithfulness while penalizing repetition. It enforces a strict one-to-one mapping $\mathcal{M}$ between $\mathcal{E}_{c,t}$ and $\mathcal{G}_{c,t}$ via the Hungarian algorithm on a thresholded similarity margin (full formulation in Appendix~\ref{app:rams_algo}). To target the \emph{paragraph bloat / information dumping} failure mode where systems pad content without adding coverage, \raaf further down-weights each generated entry by a redundancy weight $\omega(e)$ based on its maximum internal similarity to other generated entries. The Precision ($P$) and Recall ($R$) are formulated as:
\begin{equation}
\begin{aligned}
P &= \frac{1}{|\mathcal{E}_{c,t}|} \sum_{(e, g) \in \mathcal{M}} \omega(e) \cdot \mathbb{I}[\mathrm{sim}(e, g) \geq \tau_c], \\
R &= \frac{1}{|\mathcal{G}_{c,t}|} \sum_{(e, g) \in \mathcal{M}} \mathbb{I}[\mathrm{sim}(e, g) \geq \tau_c],
\end{aligned}
\end{equation}
where $\mathbb{I}[\cdot]$ is the indicator function and $\omega(e)$ is the redundancy penalty, defined as
\begin{equation}
\omega(e) = \exp\bigl(-\lambda \cdot \max_{e' \in \mathcal{E}_{c,t} \setminus \{e\}} \mathrm{sim}(e, e')\bigr),
\end{equation}
with hyperparameter $\lambda>0$. The redundancy weight $\omega(e)$ applies only to the precision term because paragraph bloat is a generation-side failure mode; the ground-truth side $\mathcal{G}_{c,t}$ is human-curated and assumed non-redundant by construction. The final \raaf is the harmonic mean of $P$ and $R$:
\begin{equation}
\raafm = \frac{2PR}{P+R}.
\end{equation}

\begin{table*}[!t]
    \centering
    \small
    \setlength{\tabcolsep}{2.5pt}
    \renewcommand{\arraystretch}{1}

    \resizebox{\textwidth}{!}{%
    \begin{tabular}{
    l| 
    c c c c c
    c c c c c
    |c 
    }
    \toprule
    \textbf{System} &
    \textbf{Edu\icon{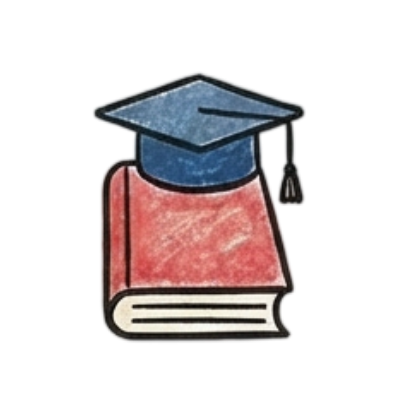}}\,$\uparrow$ & \textbf{Med\icon{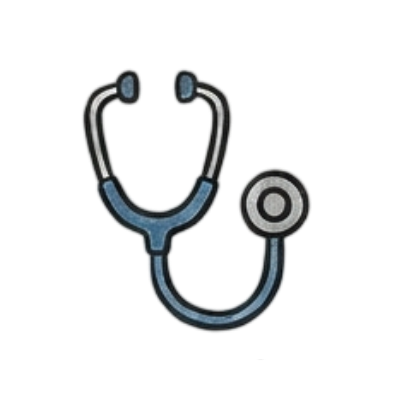}}\,$\uparrow$ & \textbf{Bio\icon{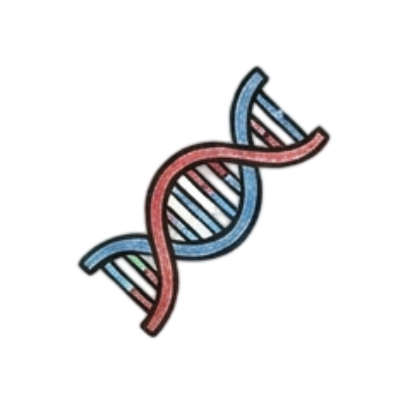}}\,$\uparrow$ & \textbf{Soc\icon{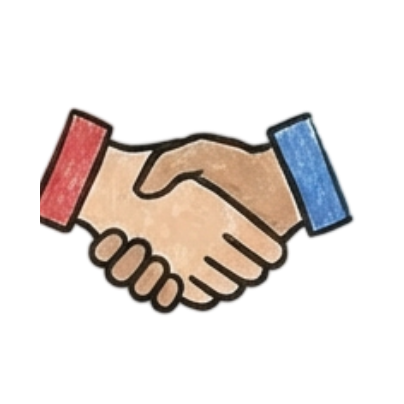}}\,$\uparrow$ & \textbf{Phys\icon{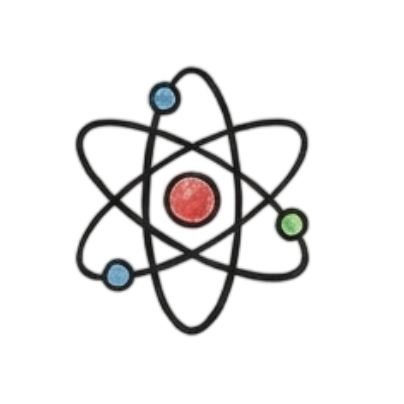}}\,$\uparrow$ &
    \textbf{Psy\icon{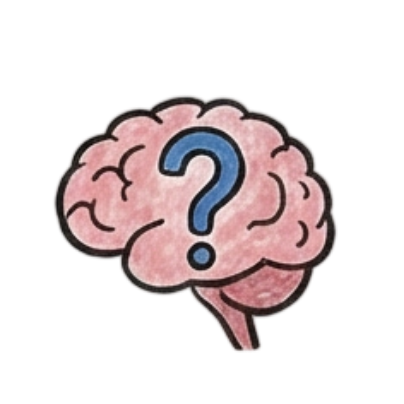}}\,$\uparrow$ & \textbf{Eng\icon{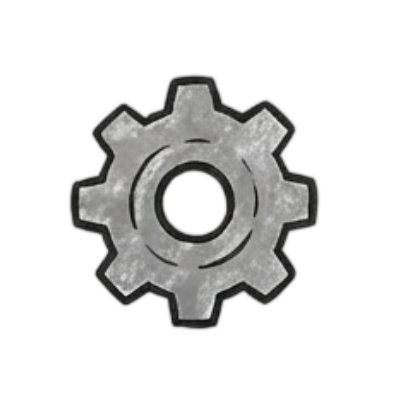}}\,$\uparrow$ & \textbf{Env\icon{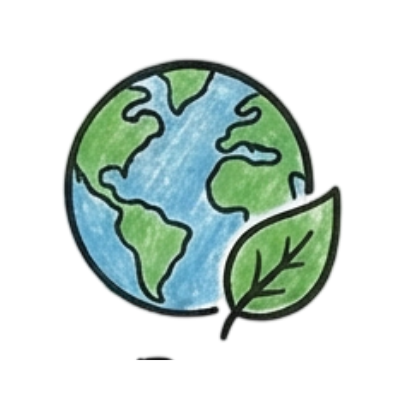}}\,$\uparrow$ & \textbf{CS\icon{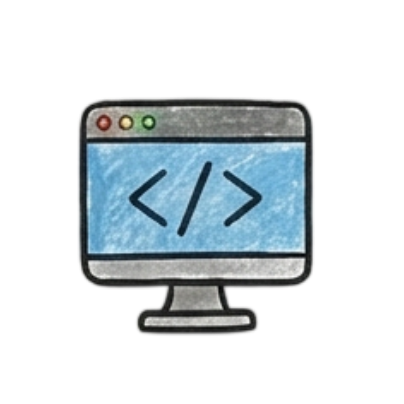}}\,$\uparrow$ & \textbf{Bus\icon{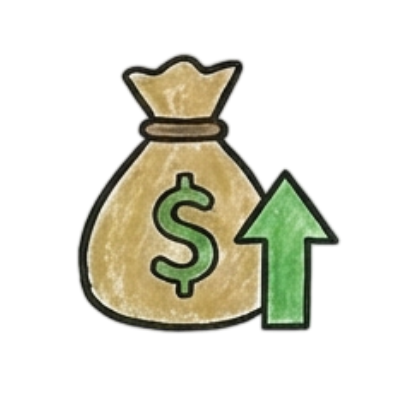}}\,$\uparrow$ &
    \textbf{Avg.}\,$\uparrow$ \\
    \midrule

    \rowcolor{slbeige}
    \multicolumn{12}{c}{\textbf{ASG Systems}} \\
    AutoSurvey           & 3.29\textcolor{green!60!black}{$_{+0.37}$} & 3.68\textcolor{green!60!black}{$_{+0.47}$} & \underline{3.63}\textcolor{green!60!black}{$_{+0.57}$} & 3.10\textcolor{green!60!black}{$_{+0.38}$} & 3.44\textcolor{green!60!black}{$_{+0.34}$} & 2.99\textcolor{green!60!black}{$_{+0.16}$} & 3.61\textcolor{green!60!black}{$_{+0.51}$} & \textbf{3.64}\textcolor{green!60!black}{$_{+0.60}$} & \underline{3.64}\textcolor{green!60!black}{$_{+0.60}$} & \underline{3.43}\textcolor{green!60!black}{$_{+0.60}$} & 3.45\textcolor{green!60!black}{$_{+0.46}$} \\
    SurveyForge          & \underline{3.33}\textcolor{green!60!black}{$_{+0.41}$} & \underline{3.84}\textcolor{green!60!black}{$_{+0.63}$} & 3.23\textcolor{green!60!black}{$_{+0.17}$} & \underline{3.23}\textcolor{green!60!black}{$_{+0.51}$} & \underline{3.63}\textcolor{green!60!black}{$_{+0.53}$} & 3.38\textcolor{green!60!black}{$_{+0.55}$} & \textbf{3.74}\textcolor{green!60!black}{$_{+0.64}$} & 3.52\textcolor{green!60!black}{$_{+0.48}$} & 3.40\textcolor{green!60!black}{$_{+0.36}$} & 3.41\textcolor{green!60!black}{$_{+0.58}$} & \underline{3.47}\textcolor{green!60!black}{$_{+0.48}$} \\
    AutoSurvey2          & 2.38\textcolor{red}{$_{-0.54}$} & 2.45\textcolor{red}{$_{-0.76}$} & 1.90\textcolor{red}{$_{-1.16}$} & 2.21\textcolor{red}{$_{-0.51}$} & 2.56\textcolor{red}{$_{-0.54}$} & 2.07\textcolor{red}{$_{-0.76}$} & 2.56\textcolor{red}{$_{-0.54}$} & 2.35\textcolor{red}{$_{-0.69}$} & 2.45\textcolor{red}{$_{-0.59}$} & 2.32\textcolor{red}{$_{-0.51}$} & 2.33\textcolor{red}{$_{-0.66}$} \\
    InteractiveSurvey    & 2.81\textcolor{red}{$_{-0.11}$} & 3.44\textcolor{green!60!black}{$_{+0.23}$} & 2.79\textcolor{red}{$_{-0.27}$} & 1.62\textcolor{red}{$_{-1.10}$} & 3.42\textcolor{green!60!black}{$_{+0.32}$} & 2.64\textcolor{red}{$_{-0.19}$} & 3.49\textcolor{green!60!black}{$_{+0.39}$} & 2.98\textcolor{red}{$_{-0.06}$} & 3.39\textcolor{green!60!black}{$_{+0.35}$} & 2.88\textcolor{green!60!black}{$_{+0.05}$} & 2.95\textcolor{red}{$_{-0.04}$} \\
    LLM$\times$MapReduce-V2     & 3.11\textcolor{green!60!black}{$_{+0.19}$} & 3.34\textcolor{green!60!black}{$_{+0.13}$} & 3.11\textcolor{green!60!black}{$_{+0.05}$} & 2.89\textcolor{green!60!black}{$_{+0.17}$} & 3.14\textcolor{green!60!black}{$_{+0.04}$} & \underline{3.47}\textcolor{green!60!black}{$_{+0.64}$} & 3.24\textcolor{green!60!black}{$_{+0.14}$} & 3.19\textcolor{green!60!black}{$_{+0.15}$} & 3.24\textcolor{green!60!black}{$_{+0.20}$} & 2.73\textcolor{red}{$_{-0.10}$} & 3.15\textcolor{green!60!black}{$_{+0.16}$} \\
    SurveyX              & 2.85\textcolor{red}{$_{-0.07}$} & 3.14\textcolor{red}{$_{-0.07}$} & 3.30\textcolor{green!60!black}{$_{+0.24}$} & 2.57\textcolor{red}{$_{-0.15}$} & 3.04\textcolor{red}{$_{-0.06}$} & 2.97\textcolor{green!60!black}{$_{+0.14}$} & 3.41\textcolor{green!60!black}{$_{+0.31}$} & 3.05\textcolor{green!60!black}{$_{+0.01}$} & 3.05\textcolor{green!60!black}{$_{+0.01}$} & 2.74\textcolor{red}{$_{-0.09}$} & 3.01\textcolor{green!60!black}{$_{+0.02}$} \\
    SciSage              & 2.84\textcolor{red}{$_{-0.08}$} & 3.06\textcolor{red}{$_{-0.15}$} & 3.04\textcolor{red}{$_{-0.02}$} & 2.59\textcolor{red}{$_{-0.13}$} & 3.00\textcolor{red}{$_{-0.10}$} & 2.63\textcolor{red}{$_{-0.20}$} & 2.65\textcolor{red}{$_{-0.45}$} & 2.77\textcolor{red}{$_{-0.27}$} & 2.77\textcolor{red}{$_{-0.27}$} & 2.75\textcolor{red}{$_{-0.08}$} & 2.81\textcolor{red}{$_{-0.18}$} \\
    \rowcolor{slbeige}
    \textbf{Avg (ASG Systems)}   & 2.94\textcolor{green!60!black}{$_{+0.02}$} & 3.28\textcolor{green!60!black}{$_{+0.07}$} & 3.00\textcolor{red}{$_{-0.06}$} & 2.60\textcolor{red}{$_{-0.12}$} & 3.18\textcolor{green!60!black}{$_{+0.08}$} & 2.88\textcolor{green!60!black}{$_{+0.05}$} & 3.24\textcolor{green!60!black}{$_{+0.14}$} & 3.07\textcolor{green!60!black}{$_{+0.03}$} & 3.13\textcolor{green!60!black}{$_{+0.09}$} & 2.89\textcolor{green!60!black}{$_{+0.06}$} & 3.02\textcolor{green!60!black}{$_{+0.03}$} \\
    \midrule

    \rowcolor{slblue}
    \multicolumn{12}{c}{\textbf{Vanilla LLMs}} \\
    Qwen3-Max             & 2.67\textcolor{red}{$_{-0.25}$} & 2.78\textcolor{red}{$_{-0.43}$} & 2.52\textcolor{red}{$_{-0.54}$} & 2.58\textcolor{red}{$_{-0.14}$} & 2.44\textcolor{red}{$_{-0.66}$} & 2.29\textcolor{red}{$_{-0.54}$} & 2.06\textcolor{red}{$_{-1.04}$} & 2.72\textcolor{red}{$_{-0.32}$} & 2.13\textcolor{red}{$_{-0.91}$} & 2.46\textcolor{red}{$_{-0.37}$} & 2.47\textcolor{red}{$_{-0.52}$} \\
    Gemini-3-Pro         & 2.28\textcolor{red}{$_{-0.64}$} & 2.46\textcolor{red}{$_{-0.75}$} & 2.75\textcolor{red}{$_{-0.31}$} & 2.46\textcolor{red}{$_{-0.26}$} & 2.41\textcolor{red}{$_{-0.69}$} & 2.15\textcolor{red}{$_{-0.68}$} & 2.46\textcolor{red}{$_{-0.64}$} & 2.40\textcolor{red}{$_{-0.64}$} & 2.60\textcolor{red}{$_{-0.44}$} & 2.01\textcolor{red}{$_{-0.82}$} & 2.40\textcolor{red}{$_{-0.59}$} \\
    \rowcolor{slblue}
    \textbf{Avg (Vanilla LLMs)} & 2.48\textcolor{red}{$_{-0.44}$} & 2.62\textcolor{red}{$_{-0.59}$} & 2.64\textcolor{red}{$_{-0.42}$} & 2.52\textcolor{red}{$_{-0.20}$} & 2.43\textcolor{red}{$_{-0.67}$} & 2.22\textcolor{red}{$_{-0.61}$} & 2.26\textcolor{red}{$_{-0.84}$} & 2.56\textcolor{red}{$_{-0.48}$} & 2.37\textcolor{red}{$_{-0.67}$} & 2.24\textcolor{red}{$_{-0.59}$} & 2.43\textcolor{red}{$_{-0.56}$} \\
    \midrule

    \rowcolor{slorange}
    \multicolumn{12}{c}{\textbf{Deep Research Agents}} \\
    Qwen DR   & 2.85\textcolor{red}{$_{-0.07}$} & 3.30\textcolor{green!60!black}{$_{+0.09}$} & 3.46\textcolor{green!60!black}{$_{+0.40}$} & 3.09\textcolor{green!60!black}{$_{+0.37}$} & 3.14\textcolor{green!60!black}{$_{+0.04}$} & 3.00\textcolor{green!60!black}{$_{+0.17}$} & 3.42\textcolor{green!60!black}{$_{+0.32}$} & 3.17\textcolor{green!60!black}{$_{+0.13}$} & 3.08\textcolor{green!60!black}{$_{+0.04}$} & 2.93\textcolor{green!60!black}{$_{+0.10}$} & 3.14\textcolor{green!60!black}{$_{+0.15}$} \\
    Gemini DR & \textbf{3.67}\textcolor{green!60!black}{$_{+0.75}$} & \textbf{3.84}\textcolor{green!60!black}{$_{+0.63}$} & \textbf{3.95}\textcolor{green!60!black}{$_{+0.89}$} & \textbf{3.53}\textcolor{green!60!black}{$_{+0.81}$} & \textbf{3.88}\textcolor{green!60!black}{$_{+0.78}$} & \textbf{3.53}\textcolor{green!60!black}{$_{+0.70}$} & 3.43\textcolor{green!60!black}{$_{+0.33}$} & \underline{3.62}\textcolor{green!60!black}{$_{+0.58}$} & \textbf{3.71}\textcolor{green!60!black}{$_{+0.67}$} & \textbf{3.45}\textcolor{green!60!black}{$_{+0.62}$} & \textbf{3.66}\textcolor{green!60!black}{$_{+0.67}$} \\
    \rowcolor{slorange}
    \textbf{Avg (Deep Research Agents)}  & 3.26\textcolor{green!60!black}{$_{+0.34}$} & 3.57\textcolor{green!60!black}{$_{+0.36}$} & 3.71\textcolor{green!60!black}{$_{+0.65}$} & 3.31\textcolor{green!60!black}{$_{+0.59}$} & 3.51\textcolor{green!60!black}{$_{+0.41}$} & 3.27\textcolor{green!60!black}{$_{+0.44}$} & 3.43\textcolor{green!60!black}{$_{+0.33}$} & 3.40\textcolor{green!60!black}{$_{+0.36}$} & 3.40\textcolor{green!60!black}{$_{+0.36}$} & 3.19\textcolor{green!60!black}{$_{+0.36}$} & 3.41\textcolor{green!60!black}{$_{+0.42}$} \\
    \midrule

    \rowcolor{gray!15}
    \textbf{Avg (All)}   & 2.92\textcolor{gray!15}{$_{+0.00}$} & 3.21\textcolor{gray!15}{$_{+0.00}$} & 3.06\textcolor{gray!15}{$_{+0.00}$} & 2.72\textcolor{gray!15}{$_{+0.00}$} & 3.10\textcolor{gray!15}{$_{+0.00}$} & 2.83\textcolor{gray!15}{$_{+0.00}$} & 3.10\textcolor{gray!15}{$_{+0.00}$} & 3.04\textcolor{gray!15}{$_{+0.00}$} & 3.04\textcolor{gray!15}{$_{+0.00}$} & 2.83\textcolor{gray!15}{$_{+0.00}$} & 2.99\textcolor{gray!15}{$_{+0.00}$} \\
    \bottomrule
    \end{tabular}
    }
    \caption{Results of discipline-aware rubric evaluation scores (1--5), best performance is in \textbf{bold}, second best is \underline{underlined}. \textbf{Avg.} denotes the macro-average of the main scores across the 10 disciplines. Subscripts (\textcolor{green!60!black}{$+$}/\textcolor{red}{$-$}) indicate each system's deviation from the per-column average across all systems (last row, \textbf{Avg (All)}).}
    \label{tab:domain_summary}
    \vspace{-6pt}
\end{table*}

\paragraph{Thresholded MaxSim (\tmax).}
With cross-topic unions $\mathcal{E}_c = \bigcup_t \mathcal{E}_{c,t}$ and $\mathcal{G}_c = \bigcup_t \mathcal{G}_{c,t}$, \tmax replaces \raaf's binary indicator with a similarity margin:
\begin{equation}
\small
\tmaxm =
\frac{1}{|\mathcal{E}_{c}|}
\sum_{e\in\mathcal{E}_{c}}
\max\!\Bigl\{0,\;
\max_{g\in\mathcal{G}_{c}}\mathrm{sim}(e,g)-\tau_c\Bigr\}.
\end{equation}
Unlike binary precision, the margin quantifies the \emph{degree} of alignment, rewarding strong matches over those that barely surpass the threshold.

\section{Experiments and Result Analysis}
We apply SurveyLens to benchmark 11 representative systems across three paradigms---\textit{ASG systems}, \textit{Vanilla LLMs}, and \textit{Deep Research Agents} (DR)---on their ability to generate discipline-appropriate surveys.
\subsection{Experiment Setup}

\paragraph{Evaluated Systems.}
We first select seven representative \textbf{ASG systems}: AutoSurvey~\cite{wang_autosurvey_2024}, SurveyForge~\cite{yan_surveyforge_2025}, AutoSurvey2~\cite{wu_autosurvey2_2025}, InteractiveSurvey~\cite{wen_interactivesurvey_2025}, LLM$\times$MapReduce-V2~\cite{wang_llmtimesmapreduce-v2_2025}, SurveyX~\cite{liang_surveyx_2025}, and SciSage~\cite{shi_scisage_2025}. 
To provide a comprehensive comparison, we additionally evaluate two \textbf{Vanilla LLMs} (Qwen3-Max~\cite{yang_qwen3_2025}, Gemini-3-Pro) alongside their corresponding \textbf{Deep Research Agents} (Qwen DR~\cite{qwen_qwen_2025}, Gemini DR~\cite{gemini_gemini_2025}).

\paragraph{Comparison Settings.}
Each system receives 100 survey topics (10 per discipline), producing $1{,}062$ valid surveys ($96.5\%$ yield; per-system breakdown and stability analysis in Appendix~\ref{app:failure_analysis}). To ensure fair comparison within open-source ASG systems, we run all of them on a unified \textit{Qwen3-30B-A3B}~\cite{yang_qwen3_2025} backbone, which also serves as our primary LLM judge; \textit{Gemini-3-Pro} is used for rubric aspect synthesis (Table~\ref{tab:aggregated_aspects}) and as a secondary cross-judge (Appendix~\ref{app:judge_robustness}), and \textit{Qwen3-Embedding-8B}~\cite{zhang_qwen3_2025} for reference-based metrics. Each (system, topic) cell is averaged over three trials; per-cell standard deviations are small (mean $0.015/0.007/0.028$ for Outline/Content/Reference; max $< 0.05$). Vanilla LLMs and DR agents are evaluated using their officially deployed configurations(Appendix~\ref{app:comparison_scope}); a leakage-controlled re-analysis is provided in Appendix~\ref{app:leakage_audit}.

\subsection{Results and Analysis}

We present our experimental results and analysis from two complementary perspectives: 
\textit{discipline-aware rubric evaluation} (\S\ref{sec:rubric_result}) and
\textit{human-reference alignment analysis} (\S\ref{sec:alignment}).
  
\subsubsection{Discipline-Aware Rubric Evaluation}
\label{sec:rubric_result}
Based on discipline-aware rubrics, we calculate scores for each ASG method at the discipline level, component level (outline, content, reference), and fine-grained aspect level, respectively.
\paragraph{\textbf{Finding 1: Deep Research Agents are the only paradigm that performs robustly across all 10 disciplines.}}

As shown in Table~\ref{tab:domain_summary}, Deep Research Agents are the only paradigm that exceeds the discipline-wise average in all ten fields, whereas every ASG system and vanilla LLM falls below the paradigm average on at least one discipline. This robustness is quantitative: DR's per-discipline scores span only $0.52$ points, whereas ASG systems swing by $0.68$ points across disciplines. Strong ASG systems do take discipline-specific top scores on fields (e.g., SurveyForge on Engineering and Medicine, AutoSurvey on Environmental Science), but no ASG or vanilla system matches DR's breadth. Vanilla LLM underperformance is sharpest on Reference, where scores collapse into a density spike at $1.0$ (Figure~\ref{fig:score_distribution}), reflecting systematic citation-quality failure. This robustness is not an artefact of paired-survey leakage: excluding the 93 leaked DR surveys shifts the rubric mean by only $0.02$ with no rank flips (Appendix~\ref{app:leakage_audit}). \textit{Implication:} for practitioners outside CS, DR agents are the safest default in the absence of discipline-specific tuning.

\paragraph{\textbf{Finding 2: Data-source quality dominates architectural sophistication.}}
Within ASG systems, top performers (AutoSurvey, SurveyForge, LLM$\times$MapReduce-V2) share curated academic databases with rigorous filtering (e.g., Scholar Navigation Agent in SurveyForge~\cite{yan_surveyforge_2025}), while underperformers (SciSage, AutoSurvey2) rely on noisier web-crawled sources or less selective retrieval pipelines. \textit{Implication:} optimize a Scholar-style, citation-aware retrieval pipeline (over generic web search) before iterating on planner or writer architectures.
\paragraph{\textbf{Finding 3: ASG systems own the skeleton, DR agents own the flesh.}}
\begin{table}[t]
    \centering
    \small
    \setlength{\tabcolsep}{6pt}
    \renewcommand{\arraystretch}{1.05}
    \begin{tabular}{l c c}
        \toprule
        \textbf{System} & \textbf{Outline\,$\uparrow$} & \textbf{Content\,$\uparrow$} \\
        \midrule
        \rowcolor{slbeige}
        \multicolumn{3}{c}{\textbf{ASG Systems}} \\
        SurveyForge          & \textbf{4.48} & 3.95 \\
        AutoSurvey           & 4.11          & 3.67 \\
        \midrule
        \rowcolor{slblue}
        \multicolumn{3}{c}{\textbf{Vanilla LLMs}} \\
        Qwen3-Max             & 2.03 & \underline{4.30} \\
        Gemini-3-Pro         & 2.30 & 3.73 \\
        \midrule
        \rowcolor{slorange}
        \multicolumn{3}{c}{\textbf{Deep Research Agents}} \\
        Qwen DR   & 2.78 & \underline{4.30} \\
        Gemini DR & \underline{4.24} & \textbf{4.31} \\
        \bottomrule
    \end{tabular}
    \vspace{-6pt}
    \caption{Outline vs.\ Content averages (1--5) for representative
    systems in each paradigm. Best
    in \textbf{bold}, second best \underline{underlined}.
    Per-discipline results in Appendix~\ref{sec:detailed_results}.}
    \label{tab:finding3_skeleton_flesh}
    \vspace{-8pt}
\end{table}

\begin{table*}[!t]
  \centering
  \scriptsize
  \setlength{\tabcolsep}{2.5pt}
  \renewcommand{\arraystretch}{1}

  \newcommand{\fsep}{\rule[-3.5pt]{0.3pt}{12pt}}
  \resizebox{\textwidth}{!}{%
  \begin{tabular}{
  l|
  cccccccccc|c
  |
  cccccccccc|c
  }
  \toprule
  \multirow{2}{*}{\textbf{System}} &
  \multicolumn{11}{c|}{\textbf{\raaf}~$\uparrow$} &
  \multicolumn{11}{c}{\textbf{\tmax}~$\uparrow$} \\
  \cmidrule(lr){2-12} \cmidrule(lr){13-23}
& \icon{figures/icons/edu.png} & \icon{figures/icons/med.png} & \icon{figures/icons/bio.png}
  & \icon{figures/icons/soc.png} & \icon{figures/icons/phy.png} & \icon{figures/icons/psy.png}
  & \icon{figures/icons/eng.png} & \icon{figures/icons/env.png} & \icon{figures/icons/cs.png}
  & \icon{figures/icons/bus.png} & \textbf{Avg.}
  & \icon{figures/icons/edu.png} & \icon{figures/icons/med.png} & \icon{figures/icons/bio.png}
  & \icon{figures/icons/soc.png} & \icon{figures/icons/phy.png} & \icon{figures/icons/psy.png}
  & \icon{figures/icons/eng.png} & \icon{figures/icons/env.png} & \icon{figures/icons/cs.png}
  & \icon{figures/icons/bus.png} & \textbf{Avg.} \\
  \midrule
 
  \rowcolor{slbeige}
  \multicolumn{23}{c}{\textbf{ASG Systems}} \\
AutoSurvey
& 0.05 & 0.06 & 0.07 & 0.05 & 0.05 & 0.06 & 0.09 & 0.06 & 0.09 & 0.05 & 0.06  
& 0.25 & 0.19 & 0.28 & 0.22 & 0.18 & 0.20 & 0.31 & 0.20 & 0.36 & 0.24 & 0.24  
\\
SurveyForge
& 0.09 & 0.11 & 0.10 & 0.08 & 0.08 & 0.10 & 0.12 & 0.11 & 0.14 & 0.08 & 0.10  
& 0.25 & 0.24 & 0.25 & 0.23 & 0.21 & 0.22 & 0.27 & 0.22 & 0.34 & 0.24 & 0.25  
\\
AutoSurvey2
& 0.12 & 0.09 & 0.01 & 0.12 & 0.07 & 0.09 & 0.12 & 0.12 & 0.14 & 0.12 & 0.10  
& 0.29 & 0.24 & 0.21 & 0.26 & 0.26 & 0.27 & 0.32 & 0.26 & 0.38 & 0.30 & 0.28  
\\
InteractiveSurvey
& 0.08 & 0.07 & 0.07 & 0.06 & 0.08 & 0.07 & 0.09 & 0.07 & 0.10 & 0.07 & 0.08  
& 0.15 & 0.15 & 0.12 & 0.14 & 0.23 & 0.11 & 0.19 & 0.10 & 0.23 & 0.13 & 0.16  
\\
LLM$\times$MapReduce-V2
& 0.14 & 0.11 & 0.11 & 0.11 & 0.06 & 0.10 & 0.10 & 0.11 & 0.12 & 0.09 & 0.11  
& 0.34 & \underline{0.36} & 0.35 & 0.21 & 0.19 & 0.25 & 0.35 & 0.25 & 0.32 & 0.22 & 0.28  
\\
SurveyX
& 0.09 & 0.10 & 0.09 & 0.10 & 0.10 & 0.09 & 0.12 & 0.11 & 0.14 & 0.10 & 0.10  
& 0.18 & 0.19 & 0.18 & 0.22 & 0.28 & 0.17 & 0.26 & 0.22 & 0.27 & 0.22 & 0.22  
\\
SciSage
& \textbf{0.18} & \textbf{0.14} & 0.16 & \textbf{0.16} & 0.10 & \underline{0.14} & \underline{0.17} & \underline{0.15} & \textbf{0.18} & \underline{0.15} & \textbf{0.15}  
& \underline{0.42} & \textbf{0.40} & \underline{0.43} & \textbf{0.34} & 0.41 & \underline{0.36} & \textbf{0.45} & \textbf{0.39} & \textbf{0.48} & \textbf{0.43} & \textbf{0.41}  
\\

  \midrule
  \rowcolor{slblue}
  \multicolumn{23}{c}{\textbf{Vanilla LLMs}} \\
Qwen3-Max
& 0.13 & 0.11 & 0.16 & 0.14 & 0.10 & 0.12 & 0.14 & 0.13 & \underline{0.17} & 0.13 & 0.13  
& 0.29 & 0.30 & 0.43 & 0.29 & \textbf{0.49} & 0.31 & 0.39 & 0.29 & \underline{0.44} & 0.32 & 0.36  
\\
Gemini-3-Pro
& 0.12 & 0.09 & \underline{0.18} & 0.11 & \underline{0.12} & 0.10 & 0.12 & 0.14 & 0.14 & 0.13 & 0.13  
& 0.32 & 0.28 & \textbf{0.50} & 0.31 & \underline{0.45} & 0.31 & 0.32 & 0.33 & 0.41 & \underline{0.39} & 0.36  
\\

  \midrule
  \rowcolor{slorange}
  \multicolumn{23}{c}{\textbf{Deep Research Agents}} \\
Qwen DR
& \underline{0.18} & 0.13 & 0.15 & 0.14 & 0.08 & 0.14 & 0.15 & 0.14 & 0.15 & \textbf{0.15} & \underline{0.14}  
& \textbf{0.44} & 0.36 & 0.37 & \underline{0.31} & 0.29 & \textbf{0.37} & \underline{0.42} & \underline{0.35} & 0.40 & 0.37 & \underline{0.37}  
\\
Gemini DR
& 0.13 & \underline{0.14} & \textbf{0.20} & \underline{0.15} & \textbf{0.13} & \textbf{0.14} & \textbf{0.17} & \textbf{0.16} & 0.15 & 0.13 & \textbf{0.15}  
& 0.23 & 0.24 & 0.38 & 0.25 & 0.38 & 0.27 & 0.31 & 0.24 & 0.26 & 0.26 & 0.28  
\\

  \bottomrule
  \end{tabular}
  }
  \vspace{-6pt}
  \caption{Human-reference alignment evaluation results in 10 disciplines, best performance is in \textbf{bold}, second best is \underline{underlined}. \raaf = Redundancy-Aware Alignment F1, \tmax = Thresholded MaxSim.}
  \vspace{-6pt}
  \label{tab:asgbench_reference_based}
  
\end{table*}

As shown in Table~\ref{tab:finding3_skeleton_flesh} and Figure~\ref{fig:heatmap} (details in Appendix~\ref{sec:detailed_results}), component-level analysis reveals a pronounced trade-off between structural rigor and content richness: ASG systems concentrate strength in the A1--3 (outline) band whereas Deep Research Agents dominate A4--8 (content), and all paradigms remain weak on A9--11 (reference). ASG systems excel at outline generation: SurveyForge attains the highest Outline average ($4.48$), ahead of every DR agent. In contrast, Deep Research Agents lead on content synthesis: Gemini DR tops Content ($4.31$, with the Content ranking unchanged after excluding leaked outputs; Appendix~\ref{app:leakage_audit}) and Qwen DR ($4.30$) is close behind, whereas their Outline averages vary widely, reflecting heterogeneous structural reliability. Two backbone-controlled sub-comparisons isolate where these advantages originate. (i) Despite running on the smaller \textit{Qwen3-30B-A3B} backbone, SurveyForge's Outline average ($4.48$) far exceeds the flagship vanilla Qwen3-Max ($2.03$) and Gemini-3-Pro ($2.30$), indicating that ASG outline strength is architectural rather than backbone-driven. (ii) At fixed backbone, Gemini DR scores $4.31$ on Content vs.\ $3.73$ for Gemini-3-Pro Vanilla, attributing the gap to the DR retrieval-and-synthesis paradigm. The same comparison on the Qwen family yields no Content gap (Qwen DR $4.30$ vs.\ Qwen3-Max Vanilla $4.30$; Appendix~\ref{sec:detailed_results}), indicating that the DR paradigm's content-synthesis benefit is family-dependent rather than universal. \textit{Implication:} the non-overlapping strengths point to an ASG-planner + DR-writer hybrid; reference quality remains a shared bottleneck neither paradigm currently solves (Appendix~\ref{app:score_distributions}).

\paragraph{\textbf{Finding 4: Paradigm ranking is not constant across disciplines.}}
\label{sec:finding4}

The paradigm rank is not constant across disciplines, and on social-science fields it \emph{reverses}: Qwen3-Max Vanilla scores $2.58$ on Sociology, beating several ASG systems (Table~\ref{tab:domain_summary}). Two complementary biases drive this. \textbf{ASG systems} show a strong STEM bias: AutoSurvey drops from $3.64$ (CS) to $2.99$ (Psychology, $-0.65$), reflecting arXiv-heavy corpora and rigid hierarchies ill-suited for theoretical frameworks. \textbf{Vanilla LLMs} go the other way: Qwen3-Max scores higher in Education ($2.67$) and Sociology than in STEM fields, where its descriptive output matches the qualitative style of those disciplines. Only \textbf{Deep Research Agents} avoid both biases, as their agentic workflows adapt retrieval and structure per discipline. \textit{Implication:} system selection must be done per-target-discipline; CS-derived rankings do not transfer, and discipline-aware evaluation is a prerequisite.

\begin{figure}[t!]
    \centering
    \includegraphics[width=\columnwidth]{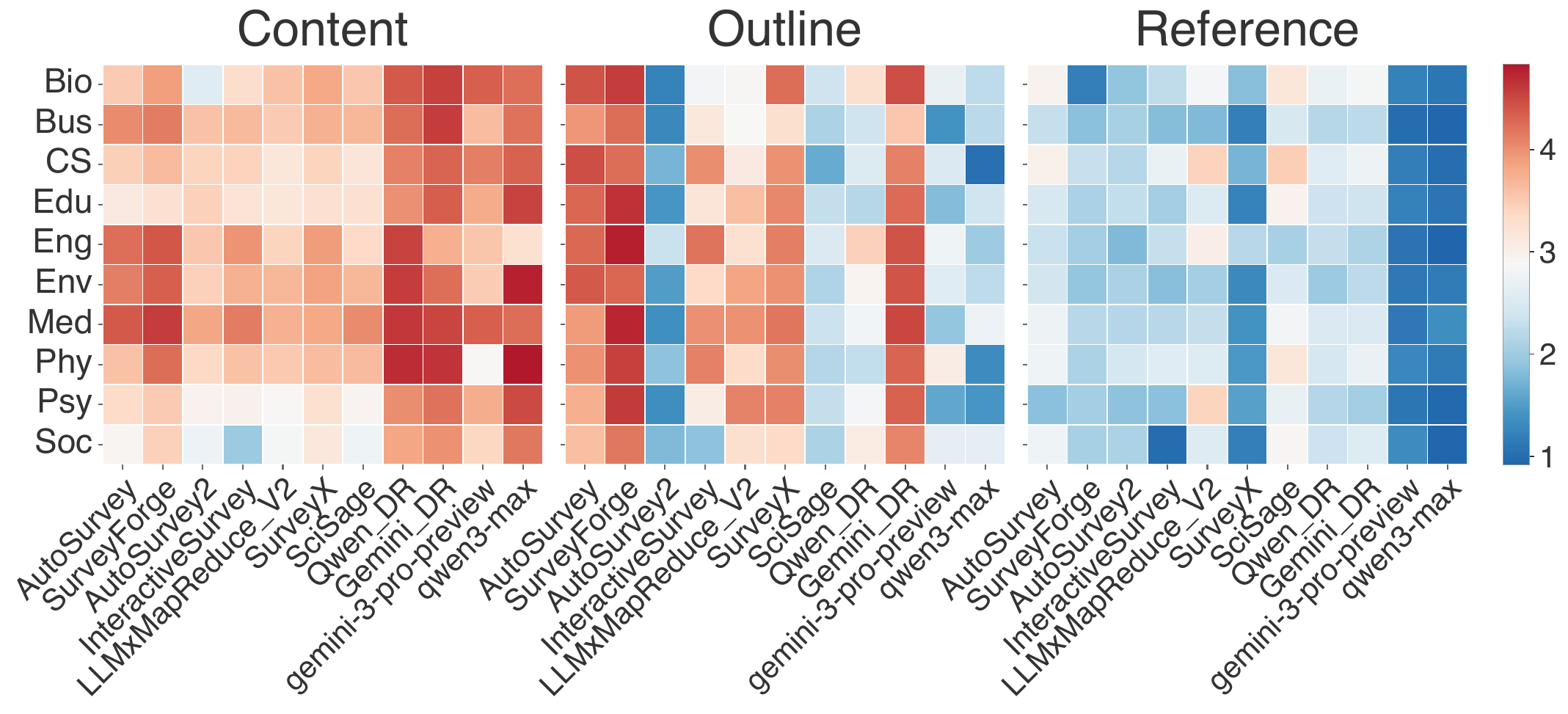}
    \caption{Aspect-level performance across the eleven sub-metrics A1--11 (grouped into Outline: A1--3; Content: A4--8; Reference: A9--11).}
    \label{fig:heatmap}
    \vspace{-8pt}
\end{figure}

\subsubsection{Human-Reference Alignment Evaluation}
\label{sec:alignment}

To complement our field-aware compliance analysis, we examine human-reference alignment through semantic similarity and structural statistics relative to paired human-written surveys.

\paragraph{\textbf{Finding 5: Models align with human surveys at the section level but fail at global organization.}}

As shown in Table~\ref{tab:asgbench_reference_based}, human-reference alignment metrics reveal a structural gap: \tmax (segment-level similarity to human text) is consistently higher than \raaf (global coverage with redundancy penalty). For example, Gemini-3-Pro averages $0.36$ vs.\ $0.13$ across the 10 disciplines, indicating local-but-not-global alignment. SciSage exemplifies this gap: high \tmax (correct factual atoms) but low discipline-aware rubric score (generic template omitting the paired survey's central contributions; full case study in Appendix~\ref{app:case_scisage}). This also explains why DR/Vanilla's high Content rubric scores should be read as local proficiency rather than survey-level competence. \textit{Implication:} future ASG and DR systems should treat global organization as an objective, not a byproduct of section-level synthesis.

\paragraph{\textbf{Finding 6: Pipeline ASG systems substitute textual bloat for substantive coverage.}}

Analysis of the ``Richness Ratio'' (Table~\ref{tab:structure_stats}) reveals a pattern of \textit{Information Dumping}. Pipeline systems exhibit extreme textual bloat (AutoSurvey: $7.11\times$ paragraphs, $4.00\times$ words; SurveyX: $4.09\times$ / $2.16\times$ vs.\ human surveys) without improving quality. Multi-modal synthesis remains weak even at best (SurveyX: $0.52\times$ images, $0.32\times$ tables; most other systems $\leq 0.05\times$), producing a ``Wall of Text'' rather than expert condensation. Appendix~\ref{app:case_autosurvey} provides a case study where AutoSurvey expands a paired human Introduction by $27\times$ via paraphrastic redundancy across eight subsections. \textit{Implication:} evaluation metrics that reward raw word or paragraph counts (e.g., naive ROUGE-recall) actively incentivize this failure mode; redundancy-aware metrics like \raaf are required.

\begin{table}[t]
    \centering
    \small
    \setlength{\tabcolsep}{6pt}
    \renewcommand{\arraystretch}{1.05}
    \begin{tabular}{l c c c c}
    \toprule
    \textbf{System} &
    \textbf{Img} & \textbf{Tab} & \textbf{Para} & \textbf{Wrd} \\
    \midrule

    \rowcolor{slbeige}
    \multicolumn{5}{c}{\textbf{ASG Systems}} \\
    AutoSurvey         & \underline{0.11} & 0.02          & \textbf{7.11}    & \textbf{4.00} \\
    SurveyX            & \textbf{0.52}    & \textbf{0.32} & \underline{4.09} & \underline{2.16} \\
    \midrule

    \rowcolor{slblue}
    \multicolumn{5}{c}{\textbf{Vanilla LLMs}} \\
    Qwen3-Max             & 0.03 & 0.04             & 0.49 & 0.21 \\
    Gemini-3-Pro         & 0.01 & 0.01             & 0.24 & 0.09 \\
    \midrule

    \rowcolor{slorange}
    \multicolumn{5}{c}{\textbf{Deep Research Agents}} \\
    Qwen DR   & 0.03 & \underline{0.26} & 0.25 & 0.22 \\
    Gemini DR & 0.02 & 0.01             & 0.27 & 0.32 \\

    \bottomrule
    \end{tabular}
    \caption{Structural ratios vs.\ human surveys (1.0 = parity). Best per column in \textbf{bold}, second best \underline{underlined}. Full results in Table~\ref{tab:structure_stats_full}.}
    \label{tab:structure_stats}
    \vspace{-6pt}
\end{table}


\subsection{Human Evaluation}
\label{sec:human_eval}

We validate the framework against human expert judgment through two complementary studies: (i) a \emph{method ablation} on the CS subset (7 PhD-level CS researchers) isolating which design choices drive alignment, and (ii) a \emph{cross-discipline generalization} test on 9 non-CS disciplines (22 PhD-level domain experts, dedicated annotation platform in Appendix~\ref{app:annotation_platform}). We anchor presentation on outline, the most reliable signal across both human and inter-LLM-judge agreement (Appendices~\ref{app:inter_dim},~\ref{app:judge_robustness}); full results are in Appendix~\ref{app:full_human_alignment}.

\begin{table}[!tbp]
\centering
\resizebox{\columnwidth}{!}{%
\begin{tabular}{@{}lcc@{}}
\toprule
\textbf{Setting} & $\rho$\,$\uparrow$ & Conc.\%\,$\uparrow$ \\
\midrule
\rowcolor{gray!15}
\multicolumn{3}{l}{\textit{(a) Method ablation on CS ($N=7$)}} \\
Holistic              & -0.13         & 43.9 \\
Discipline-Prompted   & 0.69          & 75.0 \\
Aspect (equal wt)     & 0.70          & 77.3 \\
Rubric (equal wt)     & \textbf{0.77} & 80.2 \\
Rubric + BT-domain    & \textbf{0.77} & \textbf{81.0} \\
\midrule
\rowcolor{gray!15}
\multicolumn{3}{l}{\textit{(b) Cross-discipline pooled (Rubric + BT-domain)}} \\
All non-CS ($N=22$, 9 disciplines) & \textbf{0.90} & \textbf{83.2} \\
\bottomrule
\end{tabular}}
\caption{Human evaluation: Spearman $\rho$ and pairwise concordance (Conc.\%) of LLM-judge rankings against ELO rankings. Detailed breakdown in Appendix~\ref{app:full_human_alignment}.}
\label{tab:human_alignment_outline}
\end{table}

\paragraph{Method Ablation.} 
As shown in Table~\ref{tab:human_alignment_outline}(a), holistic LLM scoring is essentially uncorrelated with human Outline judgment ($\rho = -0.13$). Discipline-Prompted ($\rho = 0.69$) helps but trails our rubric ($\rho = 0.77$, Conc.\ $= 80.2\%$); BT-domain weighting yields a modest gain ($81.0\%$).

\paragraph{Cross-discipline Generalization.} Without re-calibration, the same Rubric + BT-domain framework achieves a pooled $\rho = 0.90$ with $83.2\%$ Concordance across 9 non-CS disciplines, and \emph{every individual discipline} attains $\rho \geq 0.70$ (Table~\ref{tab:human_alignment_outline}(b)).

\paragraph{Rubric Preference.}
In a parallel criterion-recognition study, experts chose our field-aware rubric as best-fitting the work $85.3\%$ of the time (vs.\ $33.3\%$ chance; binomial $p<0.001$, full per-component breakdown in Appendix~\ref{app:expert_rubric_pref}), indicating the rubric encodes substantive disciplinary norms rather than surface cues.

\section{Conclusion}
We introduce SurveyLens, a discipline-aware benchmark for ASG with a 1{,}000-survey corpus across 10 disciplines and a dual-lens framework. Benchmarking 11 state-of-the-art systems across 3 paradigms, we reveal a structure-vs-content trade-off: Deep Research agents are the most robust default for non-CS disciplines, while specialized ASG systems remain preferable when outline rigor is critical. Reference quality also remains a universal bottleneck across all surveyed paradigms. In the long term, these discipline-aware metrics can serve as optimization objectives for ASG model alignment, enabling systems to dynamically adapt to distinct disciplinary norms.

\clearpage
\section*{Limitations}

We note four limitations of SurveyLens. (1) DR agents cite or reuse the paired human-written survey on the queried topic far more often than ASG systems do; a leakage-controlled re-analysis (Appendix~\ref{app:leakage_audit}) preserves the qualitative paradigm-level conclusions, but reference-based metrics on DR outputs should be interpreted in light of this audit. (2) Rubric scores are conditioned on a single primary judge (Qwen3-30B-A3B; Gemini-3-Pro only as a secondary cross-judge, Appendix~\ref{app:judge_robustness}) and so carry residual judge-distribution bias; this is most pronounced on Content, where both cross-judge agreement and human alignment are markedly weaker than on Outline and Reference (Appendix~\ref{app:full_human_alignment}), reflecting low intrinsic expert consensus on long-form content quality and surface-level structural cues partially driving pairwise Content judgments~\cite{krishna2023longeval}. (3) Coverage is bounded by API cost: each discipline is evaluated on $10$ topics with three repeats (variance is small, $\sigma < 0.05$), the Discipline-Prompted ablation is conducted only on the CS subset, and SurveyLens-1k is a $2020$--$2025$ snapshot that will require periodic refresh as both ASG systems and the literature evolve. (4) External validity: even with 10 disciplines, our findings should be read as a \emph{representative} rather than exhaustive characterization: performance on disciplines not covered here (e.g., humanities, law) may differ systematically, and these rankings reflect the late-2025 capability frontier with rapid LLM evolution likely to shift relative paradigm standings.

\section*{Ethics Statement}

\paragraph{Data sourcing and license.}
All papers used to construct SurveyLens-1k are obtained through the official Semantic Scholar API and used in accordance with its Terms of Service. To avoid redistributing copyrighted full-text PDFs, our public release contains only (i) publicly available bibliographic metadata, (ii) the Structured Survey Representations we derive (outline, content segments, reference titles), and (iii) URLs / DOIs that point readers back to the original sources, following the precedent of S2ORC~\cite{lo-etal-2020-s2orc}. We do not distribute the original PDFs. The released dataset and code are licensed for non-commercial academic research use only.

\paragraph{Use of human annotators.}
The expert annotations described in \S\ref{sec:human_eval} were collected from PhD-level researchers who participated voluntarily; a subset of participants received compensation at locally-competitive hourly rates, while the remainder contributed time as a collegial courtesy within the authors' professional networks. The annotation platform records no personally identifying information; the public annotations are de-identified to (anonymous expert ID, discipline) tuples (Appendix~\ref{app:annotation_platform}).

\paragraph{Use of LLM proxy for preference labels.}
Bradley-Terry weights in our rubric are learned from LLM-generated pairwise preferences rather than from large-scale human labels, as motivated in \S\ref{sec:rubric}. We disclose this proxy explicitly and validate it post-hoc against genuine expert preferences in \S\ref{sec:human_eval}; readers should interpret the weights as expert-preference-aligned via an LLM proxy, not as human-aligned.

\paragraph{Risks of misuse.}
Automatic survey generation systems produce text that is stylistically indistinguishable from human-written surveys, and our results (\S\ref{sec:rubric_result}) show that all evaluated paradigms remain weak on reference quality, including fabricated or mis-attributed citations. We intend SurveyLens as a diagnostic benchmark for system developers rather than an endorsement of using current ASG outputs as substitutes for human-authored literature reviews; downstream users should treat generated surveys, and especially their reference lists, as requiring expert verification before any scholarly use.

\bibliography{custom}

\clearpage
\appendix

\section{Evaluation Methodology Details}
\label{app:eval_methodology}

\subsection{Aspect Catalog}
\label{app:aspect_catalog}

Table~\ref{tab:aggregated_aspects} lists the universal evaluation aspects $\mathcal{A}_\mathcal{O}, \mathcal{A}_\mathcal{C}, \mathcal{A}_\mathcal{R}$ synthesized from prior survey assessment methods, before the discipline-aware expand-and-merge process described in \S\ref{sec:rubric}. These aspects serve as the seed set $\mathcal{A}_c$ in Eq.~(2) and underlie the eleven sub-metrics (A1--A11) reported in the heatmap analysis (Figure~\ref{fig:heatmap}).

\begin{table*}[htbp]
\centering
\small
\begin{tabularx}{0.85\textwidth}{@{}llX@{}}
\toprule
\textbf{Components} & \textbf{Aspects} & \textbf{Descriptions} \\
\midrule
\multirow{3}{*}{Outline} 
  & Substantive Integrity & Evaluates depth, breadth, and scholarly merit of the content coverage. \\
  & Structural Coherence & Assesses logical architecture and organizational flow of the outline. \\
  & Formal Precision & Examines the technical execution of hierarchy and presentation. \\
\midrule
\multirow{5}{*}{Content} 
  & Scope and Relevance & Evaluates breadth and alignment with the central research theme. \\
  & Structural Coherence & Assesses logical organization, transitions, and narrative consistency. \\
  & Synthesis and Integration & Measures the ability to construct cohesive frameworks. \\
  & Critical Insight and Novelty & Examines the depth of critique, original frameworks, and research gaps. \\
  & Scholarly Communication & Reviews clarity, terminology precision, and citation standards. \\
\midrule
\multirow{3}{*}{Reference} 
  & Bibliometric Comprehensiveness & Evaluates coverage extent and inclusion of seminal works. \\
  & Evidential Integrity & Assesses the the reliability of claims and correctness of attributions. \\
  & Referential Pertinence and Compliance & Examines thematic alignment and citation formatting standards. \\
\bottomrule
\end{tabularx}
\vspace{-6pt}
\caption{Assessment Aspects of Survey Papers Synthesized from Existing Studies~\cite{sun_surveybench_2025,zhao_surveyeval_2025,yan_surveyforge_2025,wang_autosurvey_2024,liang_surveyx_2025,wang_llmtimesmapreduce-v2_2025,liu_agentic_2025,nguye_surveyg_2025,zhao_surveyeval_2025,sun_surveybench_2025,guo_sgsimeval_2026}.}
\label{tab:aggregated_aspects}
\end{table*}

\subsection{\raaf Algorithm}
\label{app:rams_algo}

\begin{algorithm}[htbp]
\caption{Calculation of \raaf}
\label{alg:rams}
\begin{algorithmic}[1]
\STATE \textbf{Input:} Generated entries $\{e^{(i)}\}_{i=1}^{m}$, ground-truth entries $\{g^{(j)}\}_{j=1}^{n}$, component threshold $\tau_c$, penalty $\lambda$
\STATE \textbf{Output:} $\raafm \in [0,1]$
\STATE Compute thresholded margin matrix: $\mathbf{S}[i,j] \gets \max\bigl(0,\, \mathrm{sim}(e^{(i)}, g^{(j)}) - \tau_c\bigr)$
\STATE If $m \neq n$, pad $\mathbf{S}$ with zero rows/columns to size $\max(m,n)\times\max(m,n)$
\STATE Find optimal matching: $\pi^* \gets \arg\max_{\pi}\sum_{(i,j)\in\pi}\mathbf{S}[i,j]$ via Hungarian; retain only pairs with $i\le m$ and $j\le n$, so $|\pi^*|=\min(m,n)$
\STATE For $i=1,\dots,m$: $\omega(e^{(i)}) \gets \exp\!\bigl(-\lambda \cdot \max_{k \neq i}\, \mathrm{sim}(e^{(i)}, e^{(k)})\bigr)$ if $m>1$, else $1$
\STATE $P \gets \frac{1}{m} \sum_{(i,j) \in \pi^*} \omega(e^{(i)}) \cdot \mathbb{I}[\mathrm{sim}(e^{(i)}, g^{(j)}) \geq \tau_c]$
\STATE $R \gets \frac{1}{n} \sum_{(i,j) \in \pi^*} \mathbb{I}[\mathrm{sim}(e^{(i)}, g^{(j)}) \geq \tau_c]$
\STATE \textbf{return} $\raafm \gets 2PR/(P+R)$ if $P+R>0$, else $0$
\end{algorithmic}
\end{algorithm}

\paragraph{Hyperparameter values.}
We instantiate Algorithm~\ref{alg:rams} with cosine similarity computed over \textit{Qwen3-Embedding-8B} embeddings (indexed in ChromaDB with \texttt{hnsw:space=cosine}). The threshold is set to $\tau_c = 0.95$ for all three components $c \in \{\mathcal{O}, \mathcal{C}, \mathcal{R}\}$, and the redundancy penalty is $\lambda = 1.0$ for all components. We choose $\tau_c = 0.95$ as a conservative paraphrase filter: near-synonymous rewrites under Qwen3-Embedding-8B typically fall in the $0.93$--$0.97$ cosine range, so $0.95$ sits at the upper end of that band and admits only substantive matches while filtering shallow lexical overlap. We use $\lambda = 1.0$ as a unit-scale default, which keeps the redundancy weight on the same dimensionless scale as the similarity itself. Both values are fixed before running the main evaluation and not tuned against the human-alignment data, so reported numbers are not a function of hyperparameter search.

\subsection{Rubric Weight Learning}
\label{app:rubric_weights}

The L2 regularization coefficient in Eq.~(3) is set to $\alpha = 0.01$ for all (discipline, component) pairs. Like $\tau_c$ and $\lambda$ above, this value is fixed before running the main evaluation and not tuned against the human-alignment data.

\subsection{Scoring Example}
\label{sec:scoring_example}

Per-aspect scores $x_{i,k}$ in the rubric evaluation are obtained via LLM-based assessment. Given the discipline-specific rubric $\mathcal{K}_{d,c,a}$ and the corresponding survey component, we prompt an LLM judge to assess compliance on a 5-point scale. To ensure reliability, we adopt a structured output format requiring the judge to provide a rationale before assigning each score. An illustrative scoring example is shown below.

\begin{center}
\begin{tcolorbox}[
    colback=lightbody,
    colframe=darkheader,
    coltitle=white,
    title=\textbf{\large Scoring Example (CS, Outline)},
    boxrule=1pt,
    width=0.98\linewidth,
    left=8pt, right=8pt, top=6pt, bottom=6pt,
    fonttitle=\bfseries
]

\textbf{Aspect:} Critical Insight \\[2pt]
\textbf{Learned Weight:} $\tilde{w}_{\mathcal{O},k} = 0.60$ \\[2pt]
\textbf{Rubric Criterion:} Identifies open problems \& future directions \\[2pt]
\textbf{Rubric Description:} Articulates unresolved challenges and synthesizes limitations across methods \\[2pt]
\textbf{Normalized Score:} $\bar{x}_{i,k} = 4.2 / 5.0 = 0.84$ \\[2pt]

\hrule\vspace{2pt}

\textbf{Contribute $\mathcal{S}^{\textsc{rub}}_{d,\mathcal{O}}$:}
$5 \cdot \tilde{w}_{\mathcal{O},k} \cdot \bar{x}_{i,k} = 5 \times 0.6 \times 0.84 = \mathbf{2.52}$

\end{tcolorbox}
\end{center}

\section{SurveyLens-1k Dataset}
\label{app:dataset}

\subsection{Construction Details}
\label{app:dataset_construction}

To ensure a robust and fair evaluation across diverse academic fields, we constructed \textbf{SurveyLens-1k}. This dataset comprises \textbf{1,000 high-quality survey papers}, evenly distributed across \textbf{10 disciplines} (100 papers per field). This section details our data collection, filtering, and verification pipeline.

We used the Semantic Scholar (S2) API\footnote{\url{https://api.semanticscholar.org}} to obtain a comprehensive collection of survey papers published and corresponding metadata between 2020 and 2025. To guarantee domain diversity, we targeted ten representative disciplines: Biology, Business, Computer Science, Education, Engineering, Environmental Science, Medicine, Physics, Psychology, and Sociology.

To balance efficiency and accuracy, we restrict API queries to the "Review" document type for each discipline. Subsequently, we employ a hybrid filtering strategy to identify genuine surveys: papers with standard keywords in their titles (e.g., "survey," "review") are automatically accepted, while ambiguous entries undergo LLM-based binary classification using their titles and abstracts.

Following the initial retrieval and filtering process, we obtained an Initial Survey Pool consisting of approximately 3,000 papers. To ensure the inclusion of high-quality and influential works, we employed a citation-based selection strategy. Specifically, we ranked the papers within each discipline in descending order based on \emph{both} (i) the raw \texttt{citationCount} and (ii) the \texttt{influentialCitationCount} metric returned by the Semantic Scholar API.\footnote{Per the Semantic Scholar API documentation, \texttt{influentialCitationCount} weights citations by the influence of the citing paper, so this metric is not biased toward simply older surveys with longer accumulation windows.} Combining these two signals avoids favoring merely old, highly-cited surveys while still privileging works of demonstrated impact. From this ranked list, we selected the top 110 papers per discipline. This procedure resulted in a refined set of Candidate Surveys containing roughly 1,100 papers, which serves as the foundation for our subsequent analysis.

\begin{figure}[t]
    \centering
    \includegraphics[width=\columnwidth]{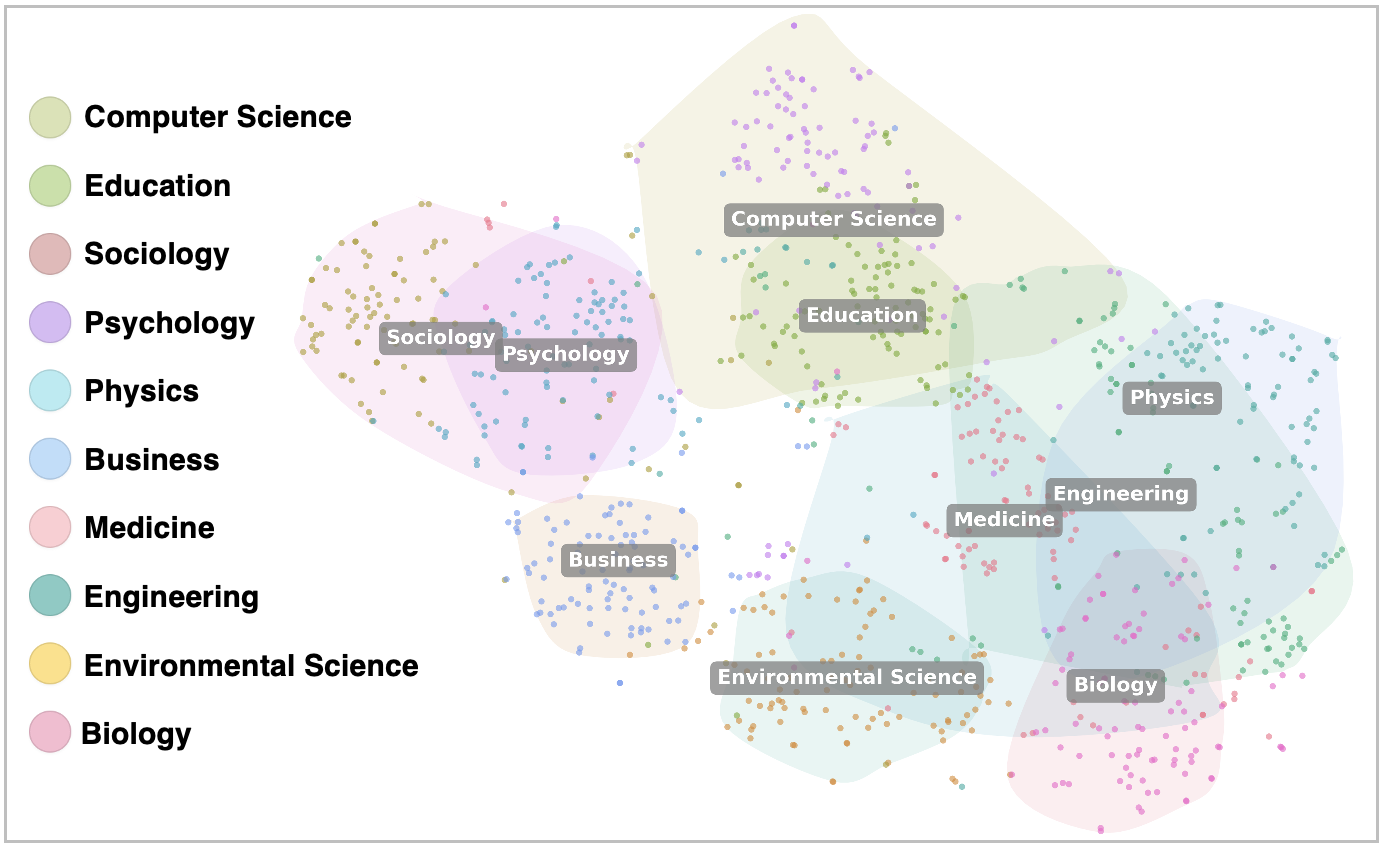}
    \vspace{-10pt}
    \caption{t-SNE visualization of survey-paper topic distribution in SurveyLens-1k, colored by discipline.}
    \label{fig:distributions}
\end{figure}

To ensure the integrity and accessibility of the dataset, we implemented a rigorous two-step verification process on the candidate surveys. First, we employed an LLM-based verification to definitively validate the paper type, filtering out any remaining non-survey articles that passed the initial screening. Subsequently, we conducted a human verification phase focused on full-text acquisition. In this step, we manually inspected the candidates to ensure PDF accessibility, discarding entries with broken links, paywalls that could not be bypassed, or corrupted files. This process yielded a finalized collection of 1,000 high-quality survey papers, constituting the SurveyLens-1k Dataset. Figure~\ref{fig:distributions} visualizes the resulting topical distribution: the 10 disciplines form well-separated clusters in topic-embedding space, indicating balanced coverage rather than concentration in any single domain.

\subsection{Per-Discipline Structural Statistics}
\label{app:dataset_stats}

Table~\ref{tab:dataset_stats} reports per-paper structural statistics averaged over the 100 human-written surveys in each of the 10 disciplines. These statistics quantify the cross-disciplinary variation summarized in \S\ref{sec:rubric} and motivate the field-aware evaluation framework: technical density (equations, words), evidence presentation (images, tables), and citation practice all vary by an order of magnitude across disciplines.

\begin{table}[t]
\centering
\resizebox{\columnwidth}{!}{%
\begin{tabular}{c|cccccccc}
\toprule
\textbf{Discipline}
& \textbf{Img} & \textbf{Tab} & \textbf{Eq} & \textbf{Para}
& \textbf{Word} & \textbf{Sent} & \textbf{Citation}
& \textbf{Reference} \\
\hline
Bio  & 4.8  & 1.7 & 71.6  & 174.9 & 13.7k & 1.5k & 113.50 & 188.97 \\
Bus  & 4.3  & 6.9 & 28.7  & 116.3 & 12.1k & 1.1k & 122.47 & 130.48 \\
CS   & 8.3  & 8.7 & 91.7  & 146.9 & 14.7k & 1.2k & 128.15 & 110.39 \\
Edu  & 4.7  & 6.7 & 35.3  & 95.8  & 10.5k & 1.0k & 98.61  & 80.71  \\
Eng  & 10.3 & 4.3 & 122.1 & 203.2 & 17.5k & 1.9k & 180.36 & 245.04 \\
Env  & 4.3  & 5.0 & 87.2  & 126.8 & 12.0k & 1.2k & 108.66 & 125.29 \\
Med  & 4.1  & 3.9 & 54.7  & 152.1 & 12.7k & 1.3k & 88.71  & 155.34 \\
Phy  & 13.3 & 3.1 & 411.4 & 256.3 & 23.4k & 2.3k & 233.03 & 261.26 \\
Psy  & 1.9  & 4.4 & 54.6  & 109.6 & 11.3k & 1.1k & 117.68 & 111.84 \\
Soc  & 1.4  & 2.4 & 25.8  & 92.9  & 9.4k  & 0.8k & 115.57 & 98.81  \\
\hline
\rowcolor{gray!15}
Avg  & 5.7  & 4.7 & 98.3  & 147.5 & 13.7k & 1.3k & 130.22 & 150.41 \\
\bottomrule
\end{tabular}%
}
\caption{Per-paper structural statistics for SurveyLens-1k, averaged over the 100 human-written surveys in each discipline. \textbf{Para} denotes paragraphs; \textbf{Word} and \textbf{Sent} are reported in thousands.}
\label{tab:dataset_stats}
\end{table}

\subsection{Structured Survey Representation Pipeline}
\label{app:ssr_pipeline}

To convert raw PDFs of both generated and human-written surveys into the Structured Survey Representation $\mathcal{S} = (\mathcal{O}, \mathcal{C}, \mathcal{R})$ used throughout our evaluation, we apply a four-stage pipeline:
\begin{enumerate}
    \item \textbf{PDF parsing.} MinerU~\cite{wang_mineru_2024} converts each PDF into structured Markdown, preserving font-size metadata, figure/table placement, and reference blocks.
    \item \textbf{Rule-based filtering.} We recover the section hierarchy from font-size metadata, discard sections with extreme length deviations from per-discipline norms, and parse the reference list using discipline-aware heuristics (e.g., venue patterns, citation styles).
    \item \textbf{LLM normalization.} An LLM pass standardizes section nesting levels across heterogeneous formats and complements the heuristic parser for reference-title extraction, particularly on entries with non-standard formatting.
    \item \textbf{Human verification.} Annotators verify the resulting outlines and reference lists against the source PDFs to establish ground truth for the human-written corpus of SurveyLens-1k.
\end{enumerate}

\subsection{Scalability and Maintenance}
\label{app:scalability}

A benchmark over 10 disciplines is only useful if it can be extended and refreshed. We summarize here the operational properties of SurveyLens that make this practical.

\paragraph{Automated construction pipeline.}
The end-to-end pipeline is fully scripted: Semantic Scholar API retrieval, hybrid keyword + LLM survey-type filtering, dual-criterion (citation count + influential citation) ranking, and structured PDF parsing into Structured Survey Representations (Appendix~\ref{app:dataset_construction}). Once configured for a new discipline, it requires only API access and standard compute, with no discipline-specific code changes.

\paragraph{Human cost per new discipline.}
Across all 10 disciplines in SurveyLens-1k, the only step that requires human attention is final verification of paper-type correctness and PDF accessibility. In our experience this consumes roughly $1$--$2$ person-days per new discipline, which is comparable to or below the human cost of adding a new task to most NLP benchmarks.

\paragraph{Discipline-aware rubric extension.}
Adding a new discipline to the rubric evaluation does not require manual rubric authoring: the discipline-aware rubric is produced by the same LLM-driven \emph{expand $\to$ merge} pipeline described in \S\ref{sec:rubric}, instantiated on the new corpus. The downstream Bradley-Terry weighting step is also automated; only the optional human-alignment validation (\S\ref{sec:human_eval}) requires domain experts.

\paragraph{Periodic refresh.}
SurveyLens-1k is currently a $2020$--$2025$ snapshot. We plan to refresh it on an annual basis by re-running the same pipeline over a rolling window (e.g., $2020$--$2026$ in the next cycle), and to continue inviting domain experts to flag rubric items that no longer reflect current disciplinary conventions. We do not commit to a fixed update cadence in this paper, but the cost of one refresh cycle is bounded by the per-discipline figure above scaled by the number of disciplines.

\section{Human Annotation and Expert Validation}
\label{app:human_validation}

\subsection{Annotation Platform and Expert Study Protocol}
\label{app:annotation_platform}

To support the cross-discipline expert validation described in \S\ref{sec:human_eval}, we built a dedicated web-based annotation platform. The platform, annotation interface screenshots, and anonymized raw annotations are all released at the project repository.\footnote{\href{https://github.com/TechnicolorGUO/SurveyLens}{\nolinkurl{https://github.com/TechnicolorGUO/SurveyLens}}}

\paragraph{Expert Recruitment.}
We recruited \textbf{30 PhD-level domain experts} in total covering all 10 disciplines: 7 CS researchers (via the method-ablation study in \S\ref{sec:human_eval}) and 23 non-CS experts (via this platform), distributed as Biology (1), Business (4), Education (4 recruited; 3 in Part~A), Engineering (3), Environmental Science (2), Medicine (3), Physics (1), Psychology (1), and Sociology (4). All experts hold (or are completing) a PhD in their listed field. Of the $23$ non-CS experts, $22$ completed both annotation parts; one (Education) completed Part~B (rubric preference) only and is therefore excluded from the Part~A pairwise pool (so Part~A has $N=22$, Part~B has $N=23$).

\paragraph{Annotation Tasks.}
Each expert completed two task types on surveys from their own discipline:
\begin{itemize}
    \item \textbf{Rubric preference (criterion recognition).} Given a survey and three rubric variants (Field-Aware, Cross-Discipline, Generic) presented in randomized order with blinded labels, the expert selects which rubric best fits the work. With $23$ Part~B participants and $3+5+3=11$ criteria per expert, the design called for $253$ annotations; we collected $252$ (Outline: $69 = 3\times 23$, Content: $114$ of $115$ — one response missing, Reference: $69 = 3\times 23$).
    \item \textbf{Pairwise system comparison.} Given two anonymized system-generated surveys on the same topic, the expert selects the better one (with a no-preference option). For each expert, the $50$ pairs are stratified by topic ($5$ pairs per topic over the $10$ topics in their discipline) and drawn from the $\binom{11}{2}=55$ system pairs; pair allocation is balanced across experts so that, pooled within a discipline, each system pair is compared in both A/B orderings (double round-robin), with partial coverage in single-expert disciplines (marked with $\dag$ in Table~\ref{tab:cross_discipline_outline}). The resulting $1{,}100$ pairwise judgments ($22$ Part~A experts $\times\, 50$ pairs each) were aggregated into per-discipline ELO rankings,\footnote{ELO uses standard online updates: initial rating $= 1500$, $K = 32$, expected score $E_A = 1/(1+10^{(R_B-R_A)/400})$, with win $= 1$ / loss $= 0$ / tie $= 0.5$. Each comparison contributes one update to each of the three per-dimension ELOs (Outline / Content / Reference), scored independently. We perform a single sweep over a randomly shuffled pair list (no iterative convergence criterion), and randomize A/B presentation order per comparison to neutralize position bias (seed $= 42$).} which we compared against SurveyLens rubric-derived rankings (Table~\ref{tab:human_alignment_outline}, panel b).
\end{itemize}
Total expert workload was approximately $14$ person-hours.

\begin{figure}[h]
    \centering
    \includegraphics[width=\columnwidth]{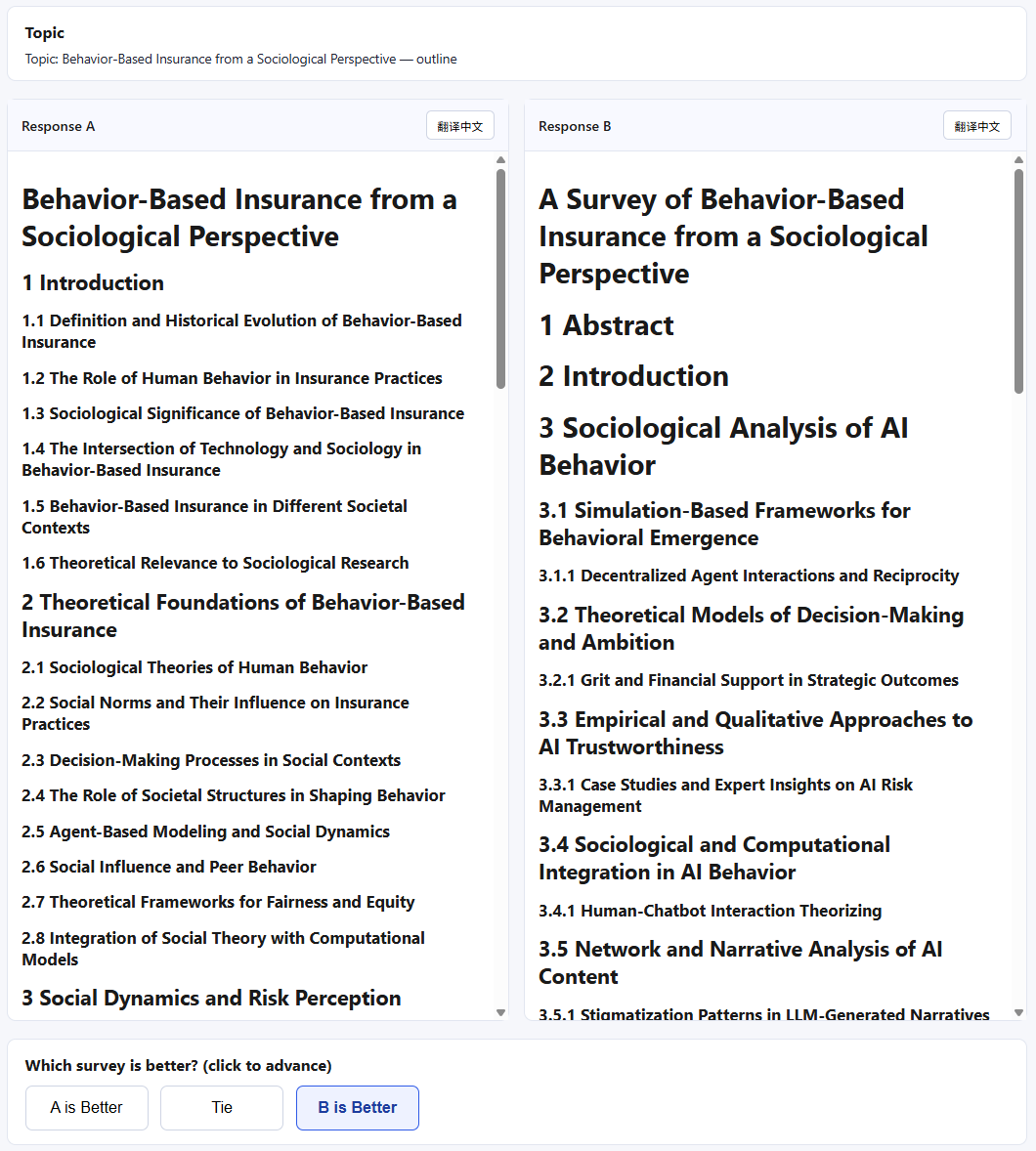}
    \caption{Part A: pairwise comparison interface. The shared survey topic is shown at the top; two anonymized system outputs are presented side-by-side under blinded labels (\emph{Response~A}, \emph{Response~B}), with model identities and any system-revealing metadata stripped. A per-pane ``translate to Chinese'' toggle is provided for non-English-native annotators. The annotator submits a judgment via one of three buttons at the bottom: \emph{A is Better}, \emph{Tie}, or \emph{B is Better}.}
    \label{fig:annotation_pairwise_ui}
\end{figure}

\begin{figure}[h]
    \centering
    \includegraphics[width=\columnwidth]{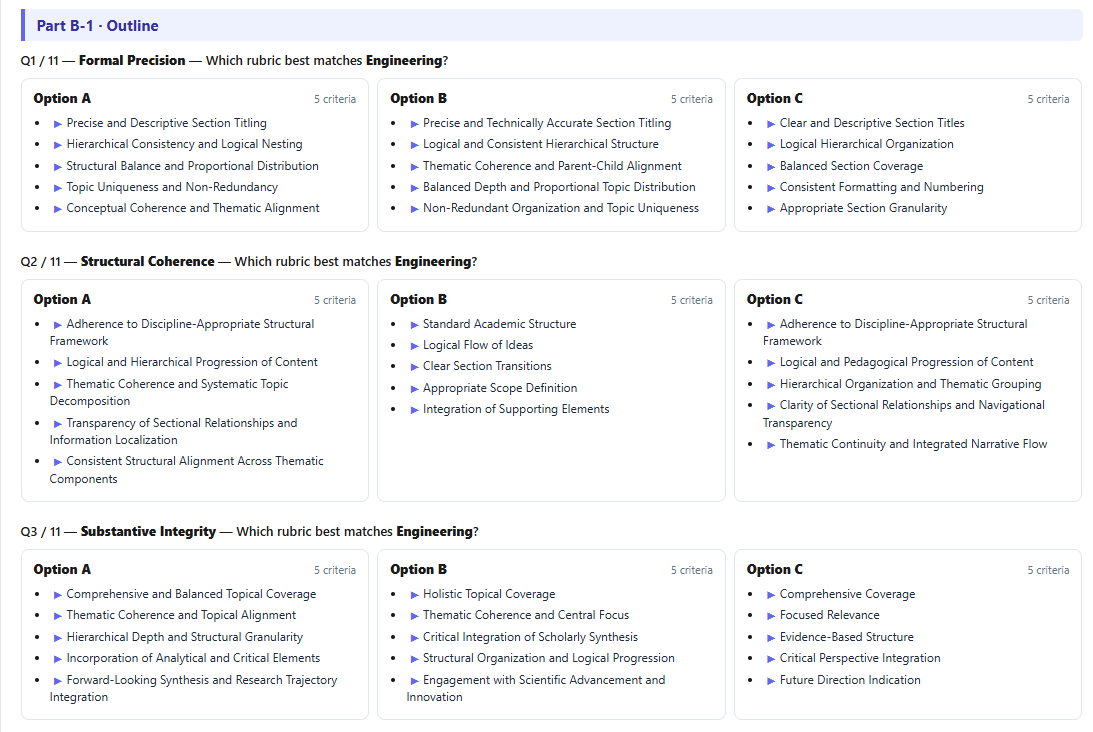}
    \caption{Part B: rubric preference interface (shown here for the Outline dimension). Each question header has the form \emph{Q\,k/N --- \textsc{Criterion} --- Which rubric best matches \textsc{Discipline}?}, naming the criterion under evaluation and the annotator's discipline (the example shown has 11 questions per session). The three rubric variants (Field-Aware, Cross-Discipline, Generic) are displayed as cards under blinded labels \emph{Option~A/B/C} in randomized order, each expanding to its 5~constituent criteria; the annotator selects the variant that best matches their discipline's academic standards.}
    \label{fig:annotation_rubric_ui}
\end{figure}

\paragraph{Platform.}
The annotation platform supports randomized presentation order, blinded labels for both rubrics and systems, session resume, and audit logging of all submitted judgments. Screenshots of the pairwise-comparison (Part A) and rubric-preference (Part B) interfaces are shown in Figures~\ref{fig:annotation_pairwise_ui} and~\ref{fig:annotation_rubric_ui}.


\paragraph{Ethics and Compensation.}
Participation was voluntary; a subset of experts received compensation at locally-competitive hourly rates, while the remainder contributed time as a collegial courtesy within the authors' academic networks. No personally identifying information is released; the public annotations are de-identified to (anonymous expert ID, discipline) tuples.

\subsection{Full Human-Alignment Results (All Three Dimensions)}
\label{app:full_human_alignment}

The main paper (\S\ref{sec:human_eval}) anchors its human-alignment validation on the Outline dimension, where both human inter-rater agreement and inter-LLM-judge agreement are highest. Here we report the full results across Outline, Content, and Reference.

\paragraph{Method ablation (CS) across all dimensions.}
Table~\ref{tab:ablation} reports Spearman $\rho$ and Pairwise Concordance for all three dimensions on the CS subset ($N=7$ experts). The rubric-based methods dominate the holistic baseline on every dimension, but the absolute alignment level is uneven: Outline and Reference reach $\rho \geq 0.47$ under Rubric + BT-domain, whereas Content alignment is substantially weaker ($\rho = 0.10$, Conc.\ $= 50.6\%$).

\begin{table}[!tbp]
\centering
\resizebox{\columnwidth}{!}{%
\begin{tabular}{@{}lcccccc@{}}
\toprule
\textbf{Setting} & \multicolumn{2}{c}{\textbf{Outline}} & \multicolumn{2}{c}{\textbf{Content}} & \multicolumn{2}{c}{\textbf{Reference}} \\
\cmidrule(lr){2-3} \cmidrule(lr){4-5} \cmidrule(lr){6-7} 
 & $\rho$ & Conc.\% & $\rho$ & Conc.\% & $\rho$ & Conc.\% \\
\midrule
Holistic & -0.13 & 43.9 & -0.17 & 35.6 & 0.18 & 56.9 \\
Discipline-Prompted & 0.69 & 75.0 & -0.21 & 41.7 & 0.37 & 61.1 \\
Aspect (equal wt) & 0.70 & 77.3 & \textbf{0.20} & \textbf{52.6} & \textbf{0.52} & 64.2 \\
Rubric (equal wt) & \textbf{0.77} & 80.2 & 0.17 & 50.6 & 0.47 & 69.3 \\
Rubric + BT-domain & \textbf{0.77} & \textbf{81.0} & 0.10 & 50.6 & 0.47 & \textbf{70.2} \\
\bottomrule
\end{tabular}}
\vspace{-6pt}
\caption{Ablation study on evaluation design choices. We report Spearman $\rho$ and Pairwise Concordance (\%) between LLM-judge scores and human preference rankings (ELO) on the CS subset. \textbf{Discipline-Prompted} injects discipline-specific instructions into the judge prompt without using our rubric.}
\label{tab:ablation}
\end{table}

\paragraph{Cross-discipline generalization (per-discipline breakdown).}
Table~\ref{tab:cross_discipline_outline} reports the per-discipline Spearman $\rho$ and Pairwise Concordance of the final framework (Rubric + BT-domain) on the Outline dimension, with per-discipline expert counts in parentheses. The pooled non-CS result ($\rho = 0.90$, Conc.\ $= 83.2\%$ over $N=22$ experts), reported in Table~\ref{tab:human_alignment_outline}(b), is supported by uniformly strong per-discipline alignment: no discipline falls below $\rho = 0.70$, and 6 of 9 disciplines reach $\rho \geq 0.90$.

\begin{table}[!tbp]
\centering
\resizebox{0.8\linewidth}{!}{
\begin{tabular}{@{}lcc@{}}
\toprule
Discipline & \textbf{$\rho$}\,$\uparrow$ & \textbf{Conc.\%}\,$\uparrow$ \\
\midrule
Biology (1)$^{\dag}$          & 0.80 & 80.0 \\
Business (4)                  & 0.70 & 78.7 \\
Education (3)                 & 0.70 & 81.2 \\
Engineering (3)               & 1.00 & 84.0 \\
Environmental Science (2)     & 0.90 & 75.3 \\
Medicine (3)                  & 1.00 & 89.3 \\
Physics (1)$^{\dag}$          & 0.90 & 84.0 \\
Psychology (1)$^{\dag}$       & 1.00 & 90.0 \\
Sociology (4)                 & 0.90 & 86.6 \\
\midrule
\textbf{All non-CS pooled ($N=22$)} & \textbf{0.90} & \textbf{83.2} \\
\bottomrule
\end{tabular}}
\caption{Per-discipline human-alignment on Outline (Rubric + BT-domain). Numbers in parentheses are expert counts; $\rho$ is computed over the $11$ systems. Cells marked $^{\dag}$ are single-annotator disciplines; the reported $\rho$ reflects rank agreement between one expert's ELO and the LLM judge, and should be read as indicative rather than statistically aggregated.}
\label{tab:cross_discipline_outline}
\end{table}

\paragraph{Why Content alignment is intrinsically harder.}
The weaker Content alignment is not specific to our rubric: holistic scoring fares worse ($\rho = -0.17$, Conc.\ $= 35.6\%$), and Discipline-Prompted does too ($\rho = -0.21$, Conc.\ $= 41.7\%$). Two independent lines of evidence indicate that this gap reflects an intrinsic property of long-form content judgment rather than a deficiency of the rubric design: (i) \emph{Human inter-dimension concordance.} Expert Content judgments concord with Outline judgments at $69.4\%$ but with Reference judgments at only $55.6\%$ (Appendix~\ref{app:inter_dim}), pointing to structural-cue interference in how humans rate content on long surveys. (ii) \emph{Inter-LLM-judge agreement.} Replicating with a second judge (Gemini-3-Pro) yields Outline and Reference $\rho \approx 0.88$ but Content $\rho = 0.34$ (Appendix~\ref{app:judge_robustness}). The same Content gap is reproduced across independent LLM judges, ruling out a rubric-side cause. We discuss mitigation strategies (content-only redaction protocols, instrument-level changes) in Limitations.

We separately note that Discipline-Prompted fares worse than Holistic on Content ($\rho = -0.21$ vs.\ $-0.17$), despite improving Outline ($\rho = 0.69$ vs.\ $-0.13$) and Reference ($\rho = 0.37$ vs.\ $0.18$). We tentatively attribute this to the discipline prompt steering the judge toward surface-level disciplinary cues (canonical structure, characteristic terminology), which aids judgment on Outline and Reference but actively crowds out substantive content assessment when applied to long-form content. The rubric, in contrast, operationalizes per-aspect criteria that constrain the judge's attention. We treat this as a hypothesis rather than a verified mechanism.

\subsection{Rubric Preference Study: Detailed Results}
\label{app:expert_rubric_pref}

Table~\ref{tab:expert_rubric_preference} reports the per-component breakdown of the 252 rubric-preference annotations summarized in \S\ref{sec:human_eval}. Across all three components, the Field-Aware rubric was chosen substantially more often than chance ($33.3\%$): Outline $94.2\%$, Content $88.6\%$, Reference $71.0\%$, pooled $85.3\%$ (binomial $p<0.001$). The relatively lower preference on Reference is consistent with citation-quality criteria being more universal across disciplines than outline conventions or content depth.

\begin{table}[!tbp]
\centering
\resizebox{\columnwidth}{!}{%
\begin{tabular}{@{}lccccc@{}}
\toprule
\textbf{Component} & \textbf{Field-Aware} & \textbf{Cross-Discipline} & \textbf{Generic} & \textbf{N} & \textbf{Binomial $p$} \\
\midrule
Outline (3 criteria)   & \textbf{94.2\%} & 5.8\%  & 0.0\%  & 69  & $<0.001$ \\
Content (5 criteria)   & \textbf{88.6\%} & 9.6\%  & 1.8\%  & 114 & $<0.001$ \\
Reference (3 criteria) & \textbf{71.0\%} & 24.6\% & 4.3\%  & 69  & $<0.001$ \\
\midrule
\textbf{Overall}       & \textbf{85.3\%} & 12.7\% & 2.0\%  & 252 & $<0.001$ \\
\bottomrule
\end{tabular}}
\vspace{-6pt}
\caption{Rubric preference study across 9 non-CS disciplines ($N=23$ Part~B experts). Experts judge which of three rubric variants best fits a given survey: \textbf{Field-Aware} (ours), \textbf{Cross-Discipline} (rubric from a different field), or \textbf{Generic} (field-agnostic). Binomial test against the $33.3\%$ chance baseline.}
\label{tab:expert_rubric_preference}
\end{table}

\subsection{Inter-dimension Concordance Analysis}
\label{app:inter_dim}

To diagnose why the Content dimension shows weaker alignment with expert judgment than Outline or Reference (\S\ref{sec:human_eval}), we examine the consistency of human pairwise preferences \textit{across} the three dimensions. For each pair of dimensions, we compute (i) the rate at which an expert who prefers system $A$ over system $B$ on one dimension also prefers $A$ over $B$ on the other (Concordance), and (ii) the Spearman $\rho$ between the per-system ELO rankings derived independently from each dimension.

As shown in Table~\ref{tab:inter_dim}, Content judgments concord with Outline judgments at $69.4\%$ ($\rho = 0.450$), substantially more than with Reference judgments ($55.6\%$, $\rho = 0.067$). If Content were judged in isolation, we would expect symmetric, near-chance concordance with both other dimensions. The observed asymmetry suggests that human raters, when making pairwise Content judgments on long-form surveys, are partially influenced by surface-level structural cues such as section titles and headings rather than purely by content quality, an effect documented in prior long-form generation studies~\cite{krishna2023longeval}. This structural-cue interference, rather than a deficiency of the rubric itself, is the most plausible explanation for the residual Content alignment gap in Table~\ref{tab:ablation}; mitigating it likely requires content-only redaction protocols or instrument-level changes to the annotation procedure, which we leave to future work.

\begin{table}[!tbp]
\centering
\resizebox{\columnwidth}{!}{%
\begin{tabular}{@{}lcc@{}}
\toprule
\textbf{Dimension Pair} & \textbf{Concordance} & \textbf{Spearman $\rho$} \\
\midrule
Outline $\times$ Content   & 69.4\% & 0.450 \\
Outline $\times$ Reference & 58.3\% & 0.267 \\
Content $\times$ Reference & 55.6\% & 0.067 \\
\bottomrule
\end{tabular}}
\vspace{-6pt}
\caption{Inter-dimension concordance among human expert pairwise judgments. For each pair of dimensions, we report the rate at which expert preferences agree on the winning system (Concordance) and the Spearman $\rho$ between the per-system ELO rankings derived from each dimension.}
\label{tab:inter_dim}
\end{table}

\section{Evaluation Robustness and Audits}
\label{app:robustness_audits}

\subsection{Multi-Judge Robustness Check}
\label{app:judge_robustness}

To assess whether our primary results are sensitive to the choice of LLM judge, we replicate the rubric evaluation with a second independent judge, \textit{Gemini-3-Pro}, on the 4 directly comparable systems (Qwen3-Max, Qwen DR, Gemini-3-Pro, and Gemini DR) across all 10 disciplines. These four systems are chosen because they form two backbone-matched Vanilla$\leftrightarrow$DR pairs, making them the most informative slice for diagnosing both inter-judge agreement and potential self-preference.

\paragraph{Inter-judge agreement.}
Table~\ref{tab:judge_robustness} reports the per-dimension Spearman $\rho$ and Pairwise Concordance between the primary judge (\textit{Qwen3-30B-A3B}) and \textit{Gemini-3-Pro}. Average agreement is high ($\rho = 0.84$, Concordance $= 88.3\%$). Outline and Reference rankings are near-identical across judges ($\rho = 0.88$ on both), whereas Content shows lower agreement ($\rho = 0.34$). This Content gap mirrors the structural-cue interference we observe in human expert judgments (Appendix~\ref{app:inter_dim}): long-form content judgments are intrinsically noisier across raters (LLM or human) than judgments anchored on outline structure or citation evidence. We note that with $N=4$ systems, Spearman $\rho$ is sensitive to single rank flips; this check should therefore be read as evidence of \emph{rank-stability under judge swap on backbone-matched pairs} rather than as a population-level inter-judge agreement statistic.

\paragraph{Absence of systematic self-preference.}
A natural concern is that each judge might favor outputs from systems sharing its model family (Qwen judge favoring Qwen3-Max/Qwen DR, Gemini judge favoring Gemini-3-Pro/Gemini DR). Examining the two pairs in question, rank shifts between the two judges for Qwen3-Max and Gemini-3-Pro are bounded by $\pm 1$ position. Crucially, the direction of any small shift is not consistent with self-preference (e.g., the Qwen judge does not uniformly rank Qwen systems higher than the Gemini judge does). We therefore find no evidence of systematic self-preference at the magnitude relevant to our paradigm-level conclusions.

\begin{table}[t]
\centering
\small
\setlength{\tabcolsep}{6pt}
\begin{tabular}{lcc}
\toprule
\textbf{Dimension} & \textbf{Inter-judge $\rho$} & \textbf{Concordance} \\
\midrule
Outline   & 0.88 & 91.7\% \\
Content   & 0.34 & 63.3\% \\
Reference & 0.88 & 90.0\% \\
\midrule
\textbf{Average} & \textbf{0.84} & \textbf{88.3\%} \\
\bottomrule
\end{tabular}
\caption{Inter-judge agreement between the primary judge (\textit{Qwen3-30B-A3B}) and a second independent judge (\textit{Gemini-3-Pro}) on 4 directly comparable systems (Qwen3-Max, Qwen DR, Gemini-3-Pro, Gemini DR) across 10 disciplines. Outline and Reference exhibit near-identical rankings; the lower Content $\rho$ mirrors the human-judge structural-cue interference reported in Appendix~\ref{app:inter_dim}.}
\label{tab:judge_robustness}
\end{table}

\subsection{Retrieval Leakage Audit Protocol}
\label{app:leakage_audit}

This appendix details a two-stage retrieval-leakage audit motivated by a concern about human-reference alignment metrics: for every generated survey, we determine whether the system reproduced or directly retrieved the paired human-written survey on the queried topic, which would inflate human-reference alignment metrics in a way that does not reflect genuine synthesis ability.

\paragraph{Stage 1: automatic candidate flagging.}
For each of the $1{,}062$ generated outputs, we (i) extract the full reference list and (ii) scan the generated body text for occurrences of the paired survey title (matched both verbatim and under a normalized form that strips punctuation, case, and version suffixes). Any output that cites the paired survey title in either location is flagged as a candidate leak.

\paragraph{Stage 2: manual verification.}
Every flagged output is then verified by hand to remove false positives such as incidental citation of a same-topic but non-paired reference, or citation of a related survey on a strictly different sub-topic. Only outputs that demonstrably ground a substantial portion of their content on the paired survey are counted as ``confirmed leakage'' in Table~\ref{tab:leakage} (panel a).

\paragraph{Coverage and per-paradigm rates.}
We apply the protocol uniformly across all $1{,}062$ outputs. Confirmed leakage rates are $2/200$ ($1.00\%$) for Vanilla LLMs, $46/665$ ($6.92\%$) for ASG systems, and $93/197$ ($47.21\%$) for Deep Research Agents. The DR category is internally split: Qwen DR $8.2\%$ vs.\ Gemini DR $85.9\%$. We report this asymmetry transparently rather than aggregate it away.

\paragraph{Leakage-excluded re-analysis.}
Table~\ref{tab:leakage} (panel b) repeats our human-reference alignment analysis with all confirmed-leaked DR outputs removed. On \raaf, the paradigm ordering DR ($0.14$) $>$ Vanilla ($0.13$) $>$ ASG ($0.10$) is preserved after exclusion, indicating that the DR advantage over ASG is not an artifact of citing the paired survey; \raaf is structurally robust to this form of leakage because its Hungarian one-to-one matching prevents repeated alignment with a single source paper. On \tmax, we note transparently that Vanilla LLMs ($0.36$) edge out leakage-excluded DR ($0.35$), with both clearly above ASG ($0.26$). We attribute the Vanilla \tmax edge to output-length effects: Vanilla outputs are short and contain little textual bloat, so their per-segment maximum similarity to the paired survey is mechanically inflated. We therefore read \raaf, which jointly penalizes redundancy and rewards coverage, as the more reliable human-reference alignment signal, on which the three-way ordering remains clean.

\begin{table}[t]
\centering
\small
\setlength{\tabcolsep}{6pt}

\textit{(a) Confirmed leakage rate by paradigm}\\[2pt]
\begin{tabular}{lcc}
\toprule
\textbf{Group} & \textbf{N} & \textbf{Leakage Rate}\,$\downarrow$ \\
\midrule
Vanilla LLMs           & 200   & $1.00\%$   \\
ASG Systems            & 665   & $6.92\%$   \\
Deep Research Agents   & 197   & $47.21\%$  \\
\midrule
\textbf{Overall}       & \textbf{1{,}062} & $\mathbf{13.29\%}$ \\
\bottomrule
\end{tabular}

\vspace{6pt}
\textit{(b) Human-reference alignment after leakage exclusion}\\[2pt]
\resizebox{\columnwidth}{!}{%
\begin{tabular}{lccc}
\toprule
\textbf{Group} & \textbf{N} & \textbf{\raaf}\,$\uparrow$ & \textbf{\tmax}\,$\uparrow$ \\
\midrule
DR (all)               & 197 & 0.15 & 0.32 \\
DR (w/o leakage)  & 104 & 0.14 & 0.35 \\
Vanilla LLM            & 200 & 0.13 & \textbf{0.36} \\
ASG                    & 665 & 0.10 & 0.26 \\
\bottomrule
\end{tabular}}

\vspace{6pt}
\textit{(c) Rubric scores after leakage exclusion (DR)}\\[2pt]
\resizebox{\columnwidth}{!}{%
\setlength{\tabcolsep}{4pt}%
\begin{tabular}{lccccc}
\toprule
\textbf{Group} & \textbf{N} & \textbf{Out.} & \textbf{Cont.} & \textbf{Ref.} & \textbf{Mean} \\
\midrule
DR (all)              & 197 & 3.53 & 4.32 & 2.43 & 3.43 \\
DR (clean)            & 104 & 3.50 & 4.29 & 2.42 & 3.40 \\
$\Delta$              &     & $-0.03$ & $-0.03$ & $-0.01$ & $-0.02$ \\
\midrule
Qwen DR (all)         &  98 & 2.76 & 4.28 & 2.40 & 3.15 \\
Qwen DR (clean)       &  90 & 2.78 & 4.26 & 2.40 & 3.15 \\
Gemini DR (all)       &  99 & 4.30 & 4.36 & 2.47 & 3.71 \\
Gemini DR (clean)$^{\dag}$ &  14 & 4.23 & 4.32 & 2.44 & 3.66 \\
\bottomrule
\end{tabular}}

\caption{Retrieval-leakage audit. (a) Confirmed leakage rates by paradigm. (b) Human-reference alignment (\raaf / \tmax) after removing leaked DR outputs. (c) Rubric scores (1--5) before and after leakage exclusion. The DR aggregate is stable ($\Delta = -0.02$ on the mean, $|\Delta| \leq 0.03$ on all three dimensions) and \emph{no per-discipline rank flips} are observed. Per-system splits are reported for completeness; $^{\dag}$Gemini DR clean ($n=14$, covering 6 of 10 disciplines) has high variance and should be read alongside the more robust DR aggregate. Gemini DR remains first on Content after exclusion ($4.32$), preserving the Finding 3 (Content) ordering.}
\label{tab:leakage}
\end{table}

\paragraph{Limitations of the audit.}
Our protocol detects leakage that surfaces as citation of, or near-verbatim reuse from, the paired survey title. It does not detect (i) paraphrased reuse without title citation, (ii) reuse of other same-topic surveys not designated as the paired reference, or (iii) reuse of underlying primary sources cited by the paired survey. A leakage-controlled evaluation protocol using post-cutoff topics is the natural next step, which we leave to future work.

\subsection{Per-System Generation Failure Analysis}
\label{app:failure_analysis}

Of the $1{,}100$ ($11 \times 100$) system-topic generation attempts, $1{,}062$ produced valid outputs (yield $96.5\%$). Table~\ref{tab:failure_analysis} breaks down the $38$ missing outputs by system and cause. Failures fall into four buckets: persistent server-side failures (SurveyX, $7$), output-handling mistakes on our side that did not affect generation itself (AutoSurvey2 Biology directory misplacement, $13$), late re-runs that are now included in the reported results (SciSage, $10$), and intermittent third-party API failures without retained call-level logs (others, $8$). None of these failure modes are correlated with specific disciplines or topics in a way that would bias the cross-discipline conclusions: missing outputs are distributed across disciplines roughly proportional to the per-system attempt count. The $8$ intermittent failures are sparsely scattered across affected systems ($\leq 2$ outputs each) and are not large enough to flip any per-discipline ranking at the resolution reported in \S\ref{sec:rubric_result}.

\begin{table}[t]
\centering
\small
\setlength{\tabcolsep}{4pt}
\begin{tabular}{llp{0.45\columnwidth}}
\toprule
\textbf{System} & \textbf{Missing} & \textbf{Cause} \\
\midrule
SurveyX        & 7  & Three retries still failed; server-side issue confirmed by authors. \\
AutoSurvey2    & 13 & Biology outputs written to a wrong directory; generation itself succeeded. \\
SciSage        & 10 & Late re-runs not in initial counts; included in current results. \\
Others         & 8  & Intermittent LLM/retrieval API failures; no call-level logs retained. \\
\midrule
\textbf{Total} & \textbf{38} & Yield rate $1{,}062 / 1{,}100 = 96.5\%$. \\
\bottomrule
\end{tabular}
\caption{Per-system generation failure breakdown. Failures are concentrated in system-level issues and show no systematic correlation with discipline or topic.}
\label{tab:failure_analysis}
\end{table}

\subsection{Comparison Scope}
\label{app:comparison_scope}

Our benchmark compares three paradigms that differ not only in algorithmic design but also in deployment form, and we structure the comparison accordingly. For the seven open-source ASG systems, we control backbone (a unified \textit{Qwen3-30B-A3B}) so that any performance gap among them isolates the architectural contribution of each pipeline. Vanilla LLMs and Deep Research agents, by contrast, are evaluated using their officially deployed configurations: their backbones are integral to the deployed product and not user-substitutable, so swapping them would no longer reflect how practitioners actually consume these systems. We therefore position the cross-paradigm results as a practitioner-oriented guide at the paradigm level. Where backbone-controlled architectural claims are made, they are supported by targeted sub-comparisons in the main text: Finding~3 (\S\ref{sec:rubric_result}) isolates the ASG outline advantage by holding the backbone fixed at Qwen3-30B-A3B (e.g., SurveyForge $4.48$ vs.\ flagship Qwen3-Max $2.03$), and the Vanilla$\leftrightarrow$DR pairs at fixed backbone family (Qwen3-Max vs.\ Qwen DR; Gemini-3-Pro vs.\ Gemini DR) isolate the DR retrieval-and-synthesis contribution. Together, these within-family controls let us attribute the headline paradigm-level findings to architecture rather than backbone capacity, while keeping the deployment-form comparison faithful to real usage.

\section{Detailed Results}

\label{sec:detailed_results}

In this section, we report the detailed performance of the evaluated systems. We first present a fine-grained visualization of aspect-wise metrics and then provide a breakdown of scores across different academic disciplines.

\subsection{Dimension-Level Grouped Comparison}
\begin{figure}[t]
    \centering
    \includegraphics[width=\columnwidth]{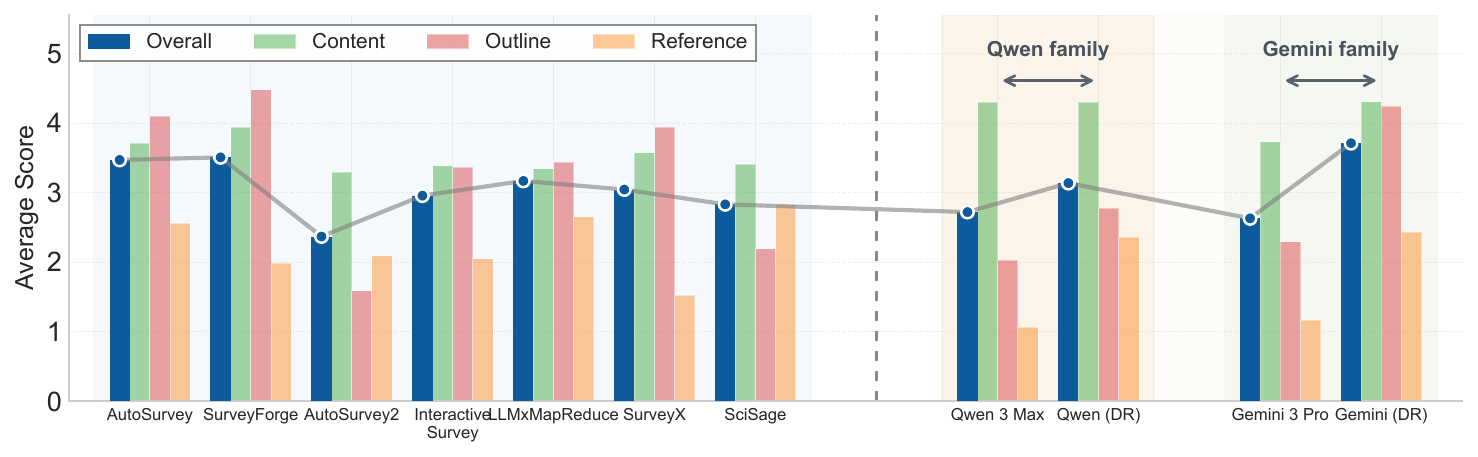}
    \vspace{-10pt}
    \caption{Dimension-level evaluation results of the 11 systems across Overall, Outline, Content, and Reference, highlighting the trade-offs between structural planning and content synthesis.}
    \label{fig:bar}
\end{figure}

Figure~\ref{fig:bar} complements the aspect-level heatmap in the main text (Figure~\ref{fig:heatmap}) by aggregating the eleven sub-metrics back into the three primary dimensions (Outline, Content, Reference) plus the Overall score, giving a side-by-side view of the 11 systems at the dimension granularity.

For reference, the aspect grouping behind both figures is as follows. The \textbf{Outline} dimension covers A1--A3, evaluating \textbf{Substantive Integrity}, \textbf{Structural Coherence}, and \textbf{Formal Precision}. The \textbf{Content} dimension spans A4--A8, assessing \textbf{Scope and Relevance}, \textbf{Structural Coherence}, \textbf{Synthesis and Integration}, \textbf{Critical Insight and Novelty}, and \textbf{Scholarly Communication}. Finally, the \textbf{Reference} dimension includes A9--A11, examining \textbf{Bibliometric Comprehensiveness}, \textbf{Evidential Integrity}, and \textbf{Referential Pertinence and Compliance}.

\subsection{Paradigm-Level Score Distributions}
\label{app:score_distributions}

\begin{figure}[t]
    \centering
    \includegraphics[width=\columnwidth]{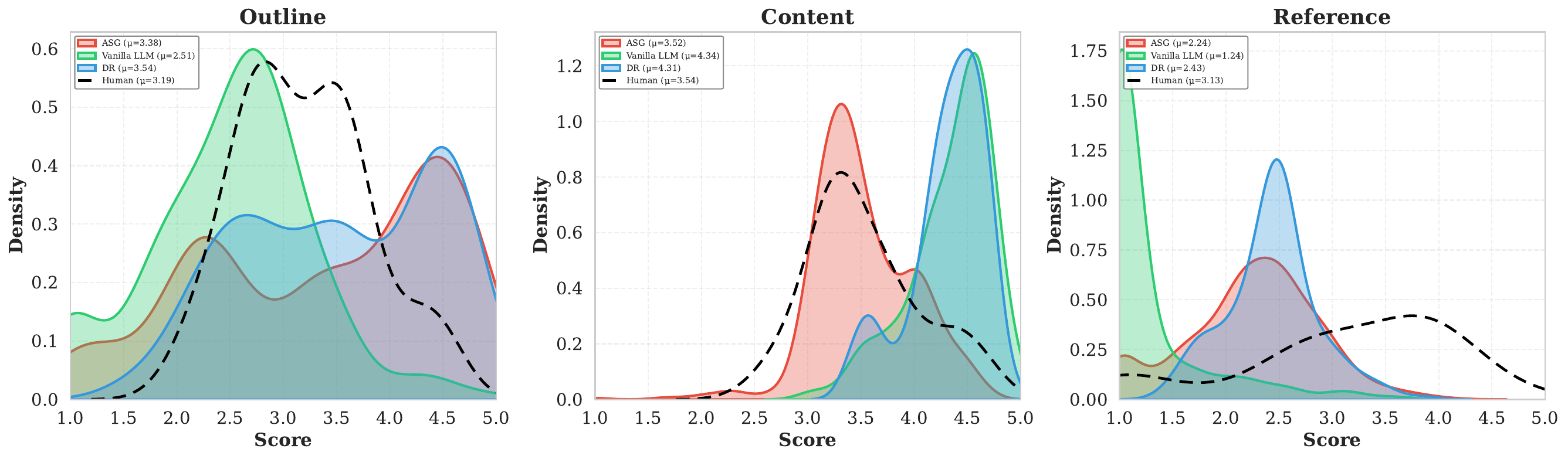}
    \vspace{-10pt}
    \caption{Paradigm-level score distributions on Outline, Content, and Reference for the three machine paradigms (ASG, Vanilla LLM, DR) and human-written surveys. The Reference panel exhibits a persistent human advantage absent from Outline and Content; Vanilla LLM exhibits a density spike at score $1.0$ on Reference; ASG Outline is bimodal at the paradigm level.}
    \label{fig:score_distribution}
\end{figure}

To complement the system-level bar chart (Figure~\ref{fig:bar}) and the aspect-level heatmap (Figure~\ref{fig:heatmap}), Figure~\ref{fig:score_distribution} reports score distributions of the three machine paradigms (ASG, Vanilla LLM, DR) alongside human-written surveys on the three primary components. Three visualisation-level observations support findings in the main text: (i) on Reference, the human distribution ($\mu = 3.13$) sits clearly above all machine paradigms (DR $\mu = 2.43$, ASG $\mu = 2.24$, Vanilla LLM $\mu = 1.24$), reinforcing Finding~3's identification of Reference as the shared bottleneck; (ii) Vanilla LLM exhibits a density spike at score $1.0$ on Reference, indicating a catastrophic citation-quality failure mode for systems without external retrieval, consistent with Finding~1's observation that vanilla LLMs consistently underperform; (iii) the ASG Outline distribution is bimodal at the paradigm level (with modes near $2.5$ and $4.5$), directly visualising the within-paradigm reliability split that Finding~2 attributes to data-source quality.

\subsection{Discipline-Specific Performance}
Tables~\ref{tab:domain_summary_outline}, \ref{tab:domain_summary_content}, and \ref{tab:domain_summary_reference} list the Reference-Free scores (1--5) for the three key dimensions across ten distinct disciplines.

\begin{itemize}
    \item \textbf{Outline Quality (Table~\ref{tab:domain_summary_outline}):} This table compares the structural quality of the generated surveys. It includes scores for ASG Systems, Vanilla LLMs, and Deep Research Agents, highlighting the best and second-best performing models in each field.

    \item \textbf{Content Quality (Table~\ref{tab:domain_summary_content}):} This table evaluates the coherence, relevance, and depth of the generated content. It allows for a comparison of how well different approaches handle domain-specific knowledge generation.

    \item \textbf{Reference Quality (Table~\ref{tab:domain_summary_reference}):} This table assesses the accuracy and relevance of the citations provided by the systems. It particularly highlights the significant performance gap between systems with retrieval capabilities (ASG Systems and Deep Research Agents) and Vanilla LLMs in terms of hallucination and citation correctness.
\end{itemize}

\clearpage
\begin{table*}[htbp]
    \centering
    \setlength{\tabcolsep}{2.5pt}
    \renewcommand{\arraystretch}{1}

    \resizebox{\textwidth}{!}{%
    \begin{tabular}{
    l| 
    c c c c c
    c c c c c
    | c
    }
    \toprule
    \textbf{System(Outline)} &
    \textbf{Edu\icon{figures/icons/edu.png}} & \textbf{Med\icon{figures/icons/med.png}} & \textbf{Bio\icon{figures/icons/bio.png}} & \textbf{Soc\icon{figures/icons/soc.png}} & \textbf{Phys\icon{figures/icons/phy.png}} &
    \textbf{Psy\icon{figures/icons/psy.png}} & \textbf{Eng\icon{figures/icons/eng.png}} & \textbf{Env\icon{figures/icons/env.png}} & \textbf{CS\icon{figures/icons/cs.png}} & \textbf{Bus\icon{figures/icons/bus.png}} & \textbf{Avg} \\
    \midrule

    \rowcolor{slbeige}
    \multicolumn{12}{c}{\textbf{ASG Systems}} \\
    AutoSurvey           & \underline{4.28} & 3.92 & 4.43 & 3.62 & 3.99 & 3.75 & 4.27 & \underline{4.38} & \textbf{4.46} & \underline{3.95} & 4.11 \\
    SurveyForge          & \textbf{4.65} & \textbf{4.73} & \textbf{4.58} & \textbf{4.16} & \textbf{4.54} & \textbf{4.60} & \textbf{4.78} & 4.29 & \underline{4.24} & \textbf{4.24} & \textbf{4.48} \\
    AutoSurvey2          & 1.43 & 1.37 & 1.23 & 1.80 & 1.88 & 1.35 & 2.33 & 1.51 & 1.73 & 1.30 & 1.59 \\
    InteractiveSurvey    & 3.18 & 4.00 & 2.81 & 1.88 & 4.11 & 3.06 & 4.20 & 3.35 & 4.01 & 3.14 & 3.37 \\
    LLM$\times$MapReduce-V2     & 3.62 & 3.99 & 2.92 & 3.28 & 3.35 & 4.09 & 3.26 & 3.85 & 3.11 & 2.88 & 3.44 \\
    SurveyX              & 4.06 & 4.19 & 4.24 & 3.36 & 4.02 & 4.10 & 4.12 & 3.98 & 3.99 & 3.28 & 3.93 \\
    SciSage              & 2.28 & 2.35 & 2.38 & 2.10 & 2.17 & 2.28 & 2.51 & 2.13 & 1.64 & 2.11 & 2.20 \\
    \midrule
    \rowcolor{slblue}
    \multicolumn{12}{c}{\textbf{Vanilla LLMs}} \\
    Qwen3-Max             & 2.40 & 2.73 & 2.24 & 2.64 & 1.33 & 1.43 & 2.00 & 2.24 & 1.03 & 2.21 & 2.03 \\
    Gemini-3-Pro         & 1.82 & 1.92 & 2.69 & 2.64 & 3.06 & 1.60 & 2.76 & 2.58 & 2.51 & 1.40 & 2.30 \\

    \midrule
    \rowcolor{slorange}
    \multicolumn{12}{c}{\textbf{Deep Research Agents}} \\
    Qwen DR   & 2.19 & 2.79 & 3.28 & 3.08 & 2.28 & 2.82 & 3.45 & 2.95 & 2.54 & 2.38 & 2.78 \\
    Gemini DR & 4.26 & \underline{4.50} & \underline{4.46} & \underline{4.08} & \underline{4.31} & \underline{4.32} & \underline{4.42} & \textbf{4.41} & 4.11 & 3.54 & \underline{4.24} \\
    \bottomrule
    \end{tabular}%
    }
    \vspace{-6pt}
    \caption{Results of discipline-aware rubric evaluation scores (1--5) in terms of Outline. Best performance is in \textbf{bold}, second best is \underline{underlined}.}
    \label{tab:domain_summary_outline}
\end{table*}

\clearpage
\begin{table*}[t]
    \centering
    \setlength{\tabcolsep}{2.5pt}
    \renewcommand{\arraystretch}{1}

    \resizebox{\textwidth}{!}{%
    \begin{tabular}{
    l| 
    c c c c c
    c c c c c
    | c
    }
    \toprule
    \textbf{System(Content)} &
    \textbf{Edu\icon{figures/icons/edu.png}} & \textbf{Med\icon{figures/icons/med.png}} & \textbf{Bio\icon{figures/icons/bio.png}} & \textbf{Soc\icon{figures/icons/soc.png}} & \textbf{Phys\icon{figures/icons/phy.png}} &
    \textbf{Psy\icon{figures/icons/psy.png}} & \textbf{Eng\icon{figures/icons/eng.png}} & \textbf{Env\icon{figures/icons/env.png}} & \textbf{CS\icon{figures/icons/cs.png}} & \textbf{Bus\icon{figures/icons/bus.png}} & \textbf{Avg} \\
    \midrule

    \rowcolor{slbeige}
    \multicolumn{12}{c}{\textbf{ASG Systems}} \\
    AutoSurvey           & 3.11 & 4.39 & 3.52 & 2.93 & 3.59 & 3.34 & 4.23 & 4.12 & 3.47 & 4.04 & 3.67 \\
    SurveyForge          & 3.25 & \underline{4.58} & 3.90 & 3.45 & 4.24 & 3.50 & \underline{4.40} & 4.34 & 3.65 & 4.14 & 3.95 \\
    AutoSurvey2          & 3.44 & 3.83 & 2.57 & 2.73 & 3.37 & 2.97 & 3.54 & 3.44 & 3.42 & 3.59 & 3.29 \\
    InteractiveSurvey    & 3.22 & 4.13 & 3.31 & 1.98 & 3.59 & 2.99 & 3.97 & 3.74 & 3.43 & 3.67 & 3.40 \\
    LLM$\times$MapReduce-V2     & 3.18 & 3.74 & 3.59 & 2.84 & 3.52 & 2.90 & 3.42 & 3.67 & 3.17 & 3.51 & 3.35 \\
    SurveyX              & 3.25 & 3.82 & 3.81 & 3.14 & 3.64 & 3.26 & 3.91 & 3.86 & 3.42 & 3.75 & 3.59 \\
    SciSage              & 3.25 & 4.04 & 3.54 & 2.75 & 3.65 & 2.94 & 3.36 & 3.67 & 3.18 & 3.68 & 3.41 \\
    \midrule
    \rowcolor{slblue}
    \multicolumn{12}{c}{\textbf{Vanilla LLMs}} \\
    Qwen3-Max             & \textbf{4.53} & 4.25 & 4.22 & \textbf{4.16} & \textbf{4.83} & \textbf{4.48} & 3.27 & \textbf{4.76} & \textbf{4.32} & 4.22 & \underline{4.30} \\
    Gemini-3-Pro         & 3.79 & 4.33 & 4.34 & 3.41 & 2.89 & 3.76 & 3.56 & 3.49 & 4.11 & 3.63 & 3.73 \\

    \midrule
    \rowcolor{slorange}
    \multicolumn{12}{c}{\textbf{Deep Research Agents}} \\
    Qwen DR   & 3.99 & \textbf{4.61} & \underline{4.38} & 3.84 & \underline{4.69} & 4.01 & \textbf{4.53} & \underline{4.57} & 4.11 & \underline{4.25} & 4.30 \\
    Gemini DR & \underline{4.35} & 4.52 & \textbf{4.55} & \underline{3.98} & 4.63 & \underline{4.22} & 3.75 & 4.22 & \underline{4.30} & \textbf{4.58} & \textbf{4.31} \\
    \bottomrule
    \end{tabular}%
    }
    \vspace{-6pt}
    \caption{Results of discipline-aware rubric evaluation scores (1--5) in terms of Content. Best performance is in \textbf{bold}, second best is \underline{underlined}.}
    \label{tab:domain_summary_content}
\end{table*}

\clearpage
\begin{table*}[t]
    \centering
    \setlength{\tabcolsep}{2.5pt}
    \renewcommand{\arraystretch}{1}

    \resizebox{\textwidth}{!}{%
    \begin{tabular}{
    l| 
    c c c c c
    c c c c c
    | c
    }
    \toprule
    \textbf{System(Reference)} &
    \textbf{Edu\icon{figures/icons/edu.png}} & \textbf{Med\icon{figures/icons/med.png}} & \textbf{Bio\icon{figures/icons/bio.png}} & \textbf{Soc\icon{figures/icons/soc.png}} & \textbf{Phys\icon{figures/icons/phy.png}} &
    \textbf{Psy\icon{figures/icons/psy.png}} & \textbf{Eng\icon{figures/icons/eng.png}} & \textbf{Env\icon{figures/icons/env.png}} & \textbf{CS\icon{figures/icons/cs.png}} & \textbf{Bus\icon{figures/icons/bus.png}} & \textbf{Avg} \\
    \midrule

    \rowcolor{slbeige}
    \multicolumn{12}{c}{\textbf{ASG Systems}} \\
    AutoSurvey           & 2.47 & 2.74 & 2.96 & 2.76 & 2.76 & 1.87 & 2.34 & 2.41 & 3.00 & 2.30 & 2.56 \\
    SurveyForge          & 2.09 & 2.20 & 1.20 & 2.08 & 2.10 & 2.04 & 2.04 & 1.93 & 2.32 & 1.86 & 1.99 \\
    AutoSurvey2          & 2.28 & 2.16 & 1.90 & 2.11 & 2.44 & 1.88 & 1.80 & 2.10 & 2.18 & 2.08 & 2.09 \\
    InteractiveSurvey    & 2.04 & 2.20 & 2.25 & 1.00 & 2.58 & 1.86 & 2.30 & 1.85 & 2.72 & 1.84 & 2.06 \\
    LLM$\times$MapReduce-V2     & 2.54 & 2.28 & 2.82 & 2.56 & 2.54 & \textbf{3.42} & \underline{3.04} & 2.05 & \underline{3.44} & 1.79 & \underline{2.65} \\
    SurveyX              & 1.24 & 1.40 & 1.84 & 1.20 & 1.47 & 1.53 & 2.20 & 1.30 & 1.74 & 1.20 & 1.51 \\
    SciSage              & \textbf{2.98} & \underline{2.80} & \textbf{3.18} & \underline{2.92} & \textbf{3.18} & 2.68 & 2.08 & \underline{2.51} & \textbf{3.48} & \underline{2.48} & \textbf{2.83} \\
    \midrule
    \rowcolor{slblue}
    \multicolumn{12}{c}{\textbf{Vanilla LLMs}} \\
    Qwen3-Max             & 1.08 & 1.36 & 1.10 & 0.94 & 1.14 & 0.96 & 0.92 & 1.16 & 1.02 & 0.94 & 1.06 \\
    Gemini-3-Pro         & 1.22 & 1.12 & 1.22 & 1.34 & 1.28 & 1.10 & 1.06 & 1.12 & 1.18 & 1.01 & 1.17 \\

    \midrule
    \rowcolor{slorange}
    \multicolumn{12}{c}{\textbf{Deep Research Agents}} \\
    Qwen DR   & 2.36 & 2.50 & 2.72 & 2.36 & 2.46 & 2.16 & 2.28 & 1.98 & 2.58 & 2.17 & 2.36 \\
    Gemini DR & 2.38 & 2.50 & 2.84 & 2.54 & 2.70 & 2.04 & 2.12 & 2.23 & 2.74 & 2.23 & 2.43 \\
    \bottomrule
    \end{tabular}%
    }
    \vspace{-6pt}
    \caption{Results of discipline-aware rubric evaluation scores (1--5) in terms of Reference. Best performance is in \textbf{bold}, second best is \underline{underlined}.}
    \label{tab:domain_summary_reference}
\end{table*}

\clearpage
\begin{table*}[!tbp]
    \centering
    \resizebox{0.85\textwidth}{!}{%
    \begin{tabular}{
    l|
    c c c c c c c c c
    }
    \toprule
    \textbf{System} &
    \textbf{Img} & \textbf{Tab} & \textbf{Eq} & \textbf{Para} & \textbf{Wrd} & \textbf{Sent} & \textbf{Cite} & \textbf{Ref} & \textbf{Char} \\
    \midrule

    \rowcolor{slbeige}
    \multicolumn{10}{c}{\textbf{ASG Systems}} \\
    AutoSurvey & 0.11 & 0.02 & 0.04 & \underline{7.11} & \underline{4.00} & \underline{2.18} & \textbf{1.03} & \textbf{0.99} & \underline{4.64} \\
    SurveyForge & 0.04 & 0.01 & 0.01 & 1.51 & 0.73 & 0.35 & 0.27 & 0.24 & 0.87 \\
    AutoSurvey2 & 0.04 & 0.00 & 0.00 & 0.18 & 0.14 & 0.07 & 0.11 & 0.10 & 0.16 \\
    InteractiveSurvey & 0.05 & 0.00 & 0.01 & 0.56 & 0.39 & 0.19 & 0.09 & 0.08 & 0.47 \\
    LLM$\times$MapReduce-V2 & 0.02 & 0.02 & 0.01 & 0.46 & 0.37 & 0.18 & 0.04 & 0.03 & 0.44 \\
    SurveyX & \textbf{0.52} & \textbf{0.32} & \textbf{0.14} & \textbf{4.09} & \textbf{2.16} & \textbf{0.85} & 0.25 & 0.22 & \textbf{2.42} \\
    SciSage & 0.01 & 0.00 & 0.00 & 0.25 & 0.13 & 0.08 & 0.07 & 0.07 & 0.16 \\
    \midrule

    \rowcolor{slblue}
    \multicolumn{10}{c}{\textbf{Vanilla LLMs}} \\
    Qwen3-Max & 0.03 & 0.04 & 0.04 & 0.49 & 0.21 & 0.20 & \underline{0.33} & 0.18 & 0.24 \\
    Gemini-3-Pro & 0.01 & 0.01 & 0.03 & 0.24 & 0.09 & 0.09 & 0.11 & 0.06 & 0.11 \\
    \midrule

    \rowcolor{slorange}
    \multicolumn{10}{c}{\textbf{Deep Research Agents}} \\
    Qwen DR & 0.03 & \underline{0.26} & 0.01 & 0.25 & 0.22 & 0.14 & 0.10 & 0.10 & 0.25 \\
    Gemini DR & \underline{0.02} & 0.01 & \underline{0.04} & 0.27 & 0.32 & 0.29 & 0.07 & \underline{0.33} & 0.29 \\

    \bottomrule
    \end{tabular}%
    }
    \vspace{-6pt}
    \caption{Full structural statistics of generated surveys relative to human-written references (1.0 = human parity). Compact four-column version used in the main text in Table~\ref{tab:structure_stats}. Highest value per column in \textbf{bold}, second highest \underline{underlined}.}
    \label{tab:structure_stats_full}
\end{table*}

\clearpage
\section{Qualitative Case Studies on ASG Failure Modes}
\label{app:case_study}

To make the abstract failure modes identified in Findings~5 and~6 concrete, we conduct two qualitative case studies on a single shared topic, \emph{``Abusive Language Training Data in Natural Language Processing''}, for which a high-quality human-written survey~\cite{vidgen_directions_2021} is available as the paired reference. We compare it against generated outputs from two representative ASG systems (\textbf{AutoSurvey}~\cite{wang_autosurvey_2024} and \textbf{SciSage}~\cite{shi_scisage_2025}), which exemplify two opposite failure patterns at the structural level.

\subsection{Case A: Information Dumping in Pipeline ASG (Finding~6)}
\label{app:case_autosurvey}

\paragraph{Setup.} Finding~6 observes that pipeline ASG systems like AutoSurvey exhibit extreme textual bloat ($7.11\times$ paragraphs, $4.00\times$ words relative to human surveys) without improving substantive coverage. We zoom into the \emph{Introduction} chapter of both surveys, which targets the same rhetorical goal: motivating the importance of the topic.

\paragraph{Quantitative comparison.} Table~\ref{tab:case_autosurvey_stats} reports per-section structural statistics. AutoSurvey expands the Introduction into 8 third-level subsections, each averaging $\approx 880$ words. The whole AutoSurvey Introduction occupies roughly \textbf{7{,}001 words / 47{,}567 characters}, whereas the paired human Introduction completes the same rhetorical function in \textbf{259 words / 1{,}851 characters}, a \textbf{$27\times$ length expansion} with no corresponding gain in coverage breadth (both introduce the same set of motivations).

\begin{table}[h]
\centering
\small
\setlength{\tabcolsep}{3pt}
\begin{tabularx}{\linewidth}{@{}Xrrr@{}}
\toprule
\textbf{Section} & \textbf{Words} & \textbf{Chars} & \textbf{Sents} \\
\midrule
Human \emph{Introduction} (full)         & 259   & 1{,}851  & 13  \\
\midrule
AutoSurvey \S1.1 \emph{Importance}        & 974   & 6{,}733  & 35  \\
AutoSurvey \S1.2 \emph{Role of Data}      & 912   & 6{,}219  & 35  \\
AutoSurvey \S1.3 \emph{Societal Impact}   & 888   & 6{,}234  & 36  \\
AutoSurvey \S1.4 \emph{Curation Challenges} & 853 & 5{,}993  & 35  \\
AutoSurvey \S1.5--1.8 (4 subsec.)     & 3{,}374 & 22{,}388 & 137 \\
\midrule
AutoSurvey \emph{Intro.} (full, 8 subsec.) & \textbf{7{,}001} & \textbf{47{,}567} & 278 \\
\bottomrule
\end{tabularx}
\vspace{-6pt}
\caption{Per-section statistics for the \emph{Introduction} chapter of the human-written survey~\cite{vidgen_directions_2021} versus AutoSurvey's generated output on the same topic.}
\label{tab:case_autosurvey_stats}
\end{table}

\paragraph{Qualitative excerpt.} The contrast becomes sharper when reading opening sentences of each. The human introduction grounds claims in specific policies, venues, and counts:

\begin{quote}\small\itshape
``\dots\ policy and regulatory developments, such as the EU's Code of Conduct on Hate, the UK Government's `Online Harms' white paper, Germany's NetzDG laws, the Public Pledge on Self-Discipline for the Chinese Internet Industry, and France's anti-hate regulation. In 2020 alone, at least three computer science venues will host workshops on online hate (TRAC and STOC at LREC, and WOAH at EMNLP), and a shared task at 2019's SemEval \dots\ reported that 800 teams downloaded the training data and 115 submitted detection systems.''~\cite{vidgen_directions_2021}
\end{quote}

\noindent AutoSurvey's \S1.1, addressing the same motivational goal, instead chains generic intensifiers without verifiable evidence:

\begin{quote}\small\itshape
``Abusive language detection in Natural Language Processing (NLP) has emerged as a critical area of research \dots\ The significance of this task cannot be overstated, as it plays a vital role in creating safer online environments and promoting the ethical development of AI systems. As online interactions become more frequent and diverse, the need to identify and mitigate abusive language has become imperative to ensure that digital spaces remain inclusive and respectful for all users \dots''
\end{quote}

\noindent The next three subsections (\S1.2 \emph{Role of Training Data}, \S1.3 \emph{Societal Impact}, \S1.4 \emph{Curation Challenges}) re-invoke near-identical hedges (\emph{``plays a critical role''}, \emph{``cannot be overstated''}, \emph{``it is essential to ensure''}) and largely overlap in claim space, producing horizontal redundancy rather than incremental coverage.

\paragraph{Takeaway.} The case directly instantiates Finding~6: AutoSurvey produces a $27\times$-longer Introduction \emph{by paraphrasing the same motivational claims across eight subsections}, rather than by expert condensation. This is also precisely the failure mode the redundancy penalty $\omega(e)$ in our \raaf metric (\S\ref{sec:alignment}) is designed to surface.

\subsection{Case B: Local Alignment without Global Structure in SciSage (Finding~5)}
\label{app:case_scisage}

\paragraph{Setup.} Finding~5 documents a systematic gap between high \tmax (local segment-level similarity to human text) and low \raaf / rubric scores (global coverage and narrative organization). SciSage on this topic provides a particularly clean illustration: its individual paragraphs reference correct facts, but its structural footprint is dominated by a generic ``introduction $\to$ literature review $\to$ conclusion'' template that omits the paired survey's central contributions.

\paragraph{Structural footprint.} Table~\ref{tab:case_scisage_outline} contrasts the second-level outline coverage. The paired human-written survey~\cite{vidgen_directions_2021} organizes its body around the \emph{full life-cycle} of an abusive-language training dataset: task formation, content properties, annotation process, annotator identity, annotation guidelines, dataset sharing, an open-science infrastructure proposal (\texttt{hatespeechdata.com}), and an evidence-based best-practice checklist. SciSage's outline collapses this seven-stage life cycle into a generic \emph{Literature Review} section with two thematic subsections.

\begin{table}[h]
\centering
\small
\begin{tabularx}{\linewidth}{@{}>{\raggedright\arraybackslash\hyphenpenalty=10000\exhyphenpenalty=10000}Xcc@{}}
\toprule
\textbf{Topical block in human survey} & \textbf{Human} & \textbf{SciSage} \\
\midrule
Task formation \& detection typology   & \checkmark & --- \\
Dataset content properties             & \checkmark & --- \\
Annotation process \& methodology      & \checkmark & partial \\
Annotator identity \& demographics     & \checkmark & --- \\
Annotation guidelines                  & \checkmark & --- \\
Dataset sharing \& open science        & \checkmark & --- \\
Best-practice recommendations          & \checkmark & --- \\
\midrule
Total L1 chapters                      & 8          & 4 \\
Total L2 subsections                   & 17         & 4 \\
\bottomrule
\end{tabularx}
\vspace{-6pt}
\caption{Outline-level coverage comparison on the same topic. SciSage retains a generic survey template, fully omitting six of the paired survey's seven central contributions and only partially covering the seventh (annotation process).}
\label{tab:case_scisage_outline}
\end{table}

\paragraph{Local fidelity is preserved.} Despite this structural collapse, individual SciSage paragraphs are factually grounded and cite the same evidence that appears in the paired survey. For example, the \emph{Recent Advancements} subsection reads:

\begin{quote}\small\itshape
``A systematic review of 63 publicly available datasets highlights the difficulties in producing large, diverse, and theoretically informed collections while minimizing biases \dots\ specialized datasets like L-HSAB \dots\ emphasize rigorous data collection procedures and annotation guidelines \dots\ inter-annotator agreement metrics like Cohen's Kappa and Krippendorff's alpha to validate annotation quality.''
\end{quote}

\noindent All three referenced facts (63-dataset systematic review, L-HSAB, Cohen's $\kappa$ / Krippendorff's $\alpha$) align with the paired human-written paper, which explains why such outputs accumulate non-trivial token-level overlap (\tmax) with the human reference.

\paragraph{Global narrative is missing.} However, this local fidelity does not survive aggregation. SciSage's two thematic subsections (\emph{Recent Advancements}, \emph{Current State of Research}) collapse the paired paper's distinct chapters on (i) \emph{who annotated} the data, (ii) \emph{how the annotation guidelines treat irony, calumny, and intent}, (iii) \emph{how datasets degrade under platform Terms-of-Service constraints}, and (iv) \emph{the proposed data-trust infrastructure for sharing} into a single undifferentiated ``advances and challenges'' block. The paired survey's central proposals (the \texttt{hatespeechdata.com} repository, the four-stage best-practice checklist, and the open-science / data-trust framework) have no corresponding section in SciSage's output.

\paragraph{Takeaway.} The case operationalises the \emph{local-alignment / global-structure gap} flagged in Finding~5: SciSage retrieves correct factual atoms (high \tmax), but the discipline-specific narrative scaffold that turns those atoms into a usable survey is replaced by a one-size-fits-all template (low \raaf, low rubric Outline score). Together, Cases~A and~B show that ASG failure modes are not symmetric: \textbf{AutoSurvey over-produces text without adding coverage}, whereas \textbf{SciSage retrieves correct facts without producing the right structure}, two opposite manifestations of the same underlying gap between local segment-level alignment and global disciplinary organization.

\clearpage
\section{Prompts}

\label{sec:appendix_prompts}

To ensure reproducibility and provide a clear understanding of our framework, we detail the specific prompts used in our experiments. These prompts were designed to guide the LLM effectively across different stages of the pipeline. They are categorized into three main functional phases:

\begin{itemize}
    \item \textbf{Criteria Construction Phase:} This phase focuses on establishing comprehensive evaluation standards. Figures~\ref{fig:prompt_aggregate}, \ref{fig:prompt_expand}, and \ref{fig:prompt_merge} illustrate the prompts used to aggregate high-level aspects, expand them into detailed criteria based on survey content, and merge criteria from various sources, respectively. This process ensures that the resulting criteria are both diverse and specific to the domain.
    
    \item \textbf{Evaluation Phase:} We adopt a modular strategy to evaluate the quality of the surveys. We present the prompts designed for the per-aspect evaluation of the survey's \textbf{outline} (Figure~\ref{fig:prompt_eval_outline_peraspect}), \textbf{content} (Figure~\ref{fig:prompt_eval_content_peraspect}), and \textbf{references} (Figure~\ref{fig:prompt_eval_reference_peraspect}). Evaluating these components independently helps minimize interference and ensures a more accurate assessment.
    
    \item \textbf{Generation Phase:} 
    Figure~\ref{fig:prompt_batch} presents the prompt used for the automated generation of systematic literature reviews. This prompt guides the model to synthesize the collected references into a complete paper based on the topic provided by the user.
\end{itemize}

\begin{figure*}[htbp]
    \centering
    
    \begin{promptbox}{Aggregate Aspects Prompt }
        
        \textbf{System prompt:} You are an academic evaluation expert. I need you to aggregate the following evaluation criteria for \{category\_name\} into \{target\_count\} universal, highly aggregated aspects.
        
        \vspace{0.5em}
        \textbf{Current aspects list:}
        
        \{formatted\_aspects\}
        
        \vspace{0.5em}
        \textbf{Requirements:}
        \begin{itemize}
            \item Aggregate these \{len(aspects\_list)\} aspects into \{target\_count\} universal aspects
            \item Each aggregated aspect should:
            \begin{itemize}
                \item Be a highly abstract, universal concept
                \item Cover multiple related original aspects
                \item Have a concise and generalizable name
                \item Have clear descriptions that reflect its universal nature
            \end{itemize}
            \item Maintain academic rigor, ensuring the aggregated aspects comprehensively cover the core content of the original aspects
            \item The number of returned aspects must be exactly \{target\_count\}
        \end{itemize}
        
        \vspace{0.5em}
        \textbf{Please return the result strictly in the following JSON array format, without any additional text:}
        
        \begin{tcolorbox}[
            colback=white,
            colframe=gray!30,
            boxrule=0.5pt,
            sharp corners,
            top=2pt, bottom=2pt
        ]
        \texttt{[} \\
        \texttt{  \{"Aspect Name 1": "Detailed description 1"\},} \\
        \texttt{  \{"Aspect Name 2": "Detailed description 2"\},} \\
        \texttt{  ...} \\
        \texttt{]}
        \end{tcolorbox}
        
        \vspace{0.5em}
        \textbf{Note:} Return only the JSON array, without any markdown code block markers or other text.
        
    \end{promptbox}
    \vspace{-10pt}
    \caption{Prompt used to aggregate multiple evaluation aspects into a smaller set of universal, highly aggregated aspects. Used in the \texttt{call\_llm\_for\_aggregation} function.}
    \label{fig:prompt_aggregate}
    
\end{figure*}

\begin{figure*}[htbp]
    \centering
    
    \begin{promptbox}{Expand Aspects Prompt }
        
        \textbf{System prompt:} You are an academic survey expert. Now you are refining the writing guideline of survey writing.
        
        \vspace{0.5em}
        \textbf{Given:}
        \begin{enumerate}
            \item Component Type: \{component\_name\} (outline/content/reference)
            \item Component Content: The actual \{component\_name\} from a survey paper
            \item Aspect: \{aspect\_name\} - \{aspect\_description\}
            \item Target Number: \{n\}
        \end{enumerate}
        
        \vspace{0.5em}
        \textbf{Task:}
        
        Expand the given aspect "\{aspect\_name\}" into \{n\} specific, detailed criteria. Each criterion should:
        \begin{itemize}
            \item Provide a clear, actionable explanation of what "\{aspect\_name\}" means in the context of \{component\_name\}
            \item Include a concrete example from the provided component content that demonstrates this criterion
            \item The example should be a summary/synthesis of relevant parts from the component content, NOT a direct verbatim quote
            \item Be specific enough to guide evaluation of survey \{component\_name\}
        \end{itemize}
        
        \vspace{0.5em}
        \textbf{Component Content:}
        
        \{component\_content\}
        
        \vspace{0.5em}
        \textbf{Aspect to Expand:}
        
        Name: \{aspect\_name\} \\
        Description: \{aspect\_description\}
        
        \vspace{0.5em}
        \textbf{Please return the result in the following JSON format:}
        
        \begin{tcolorbox}[
            colback=white,
            colframe=gray!30,
            boxrule=0.5pt,
            sharp corners,
            top=2pt, bottom=2pt
        ]
        \texttt{\{} \\
        \texttt{  "aspect\_name": "\{aspect\_name\}",} \\
        \texttt{  "expanded\_criteria": [} \\
        \texttt{    \{} \\
        \texttt{      "criterion\_name": "Specific criterion name 1",} \\
        \texttt{      "description": "Detailed explanation of what this criterion means",} \\
        \texttt{      "example": "Summarized example from the component content that demonstrates this criterion"} \\
        \texttt{    \},} \\
        \texttt{    \{} \\
        \texttt{      "criterion\_name": "Specific criterion name 2",} \\
        \texttt{      "description": "Detailed explanation of what this criterion means",} \\
        \texttt{      "example": "Summarized example from the component content that demonstrates this criterion"} \\
        \texttt{    \},} \\
        \texttt{    ...} \\
        \texttt{  ]} \\
        \texttt{\}}
        \end{tcolorbox}
        
        \vspace{0.5em}
        \textbf{Note:} Return only the JSON object, without any markdown code block markers or additional text.
        
    \end{promptbox}
    \vspace{-10pt}
    \caption{Prompt used to expand a single aspect into detailed criteria using survey content. Each criterion includes description and example. Defined as \texttt{EXPAND\_ASPECT\_PROMPT}.}
    \label{fig:prompt_expand}
    
\end{figure*}

\begin{figure*}[htbp]
    \centering
    
    \begin{promptbox}{Merge Aspect Criteria Prompt}
        
        \textbf{System prompt:} You are an academic survey expert. Now you are merging multiple expanded criteria from the same aspect across multiple survey papers into a consolidated set.
        
        \vspace{0.5em}
        \textbf{Given:}
        \begin{enumerate}
            \item Aspect Name: \{aspect\_name\}
            \item Component Type: \{component\_name\} (outline/content/reference)
            \item All Expanded Criteria: A list of criteria from the same aspect "\{aspect\_name\}" collected from multiple survey papers
            \item Target Number: \{n\}
        \end{enumerate}
        
        \vspace{0.5em}
        \textbf{Task:}
        
        Merge all the given expanded criteria into \{n\} comprehensive, consolidated criteria. Each merged criterion should:
        \begin{itemize}
            \item Synthesize related criteria from different survey papers into a unified, coherent criterion
            \item Preserve the essential meaning and evaluation focus of the original criteria
            \item Be general enough to apply across different survey papers while remaining specific enough to guide evaluation
            \item Avoid redundancy while maintaining comprehensiveness
            \item Each merged criterion should have a clear, distinct focus
            \item Extract common patterns and best practices from the examples
        \end{itemize}
        
        \vspace{0.5em}
        \textbf{All Expanded Criteria:}
        
        \{all\_criteria\_json\}
        
        \vspace{0.5em}
        \textbf{Please return the result in the following JSON format:}
        
        \begin{tcolorbox}[
            colback=white,
            colframe=gray!30,
            boxrule=0.5pt,
            sharp corners,
            top=2pt, bottom=2pt
        ]
        \texttt{\{} \\
        \texttt{  "aspect\_name": "\{aspect\_name\}",} \\
        \texttt{  "component\_name": "\{component\_name\}",} \\
        \texttt{  "merged\_expanded\_criteria": [} \\
        \texttt{    \{} \\
        \texttt{      "criterion\_name": "Consolidated criterion name 1",} \\
        \texttt{      "description": "Comprehensive description that synthesizes related criteria",} \\
        \texttt{      "example": "Highly synthesized and aggregated example that demonstrates universally applicable criteria"} \\
        \texttt{    \},} \\
        \texttt{    \{} \\
        \texttt{      "criterion\_name": "Consolidated criterion name 2",} \\
        \texttt{      "description": "Comprehensive description that synthesizes related criteria",} \\
        \texttt{      "example": "Highly synthesized and aggregated example that demonstrates universally applicable criteria"} \\
        \texttt{    \},} \\
        \texttt{    ...} \\
        \texttt{  ]} \\
        \texttt{\}}
        \end{tcolorbox}
        
        \vspace{0.5em}
        \textbf{Note:} Return only the JSON object, without any markdown code block markers or additional text.
        
    \end{promptbox}
    \vspace{-10pt}
    \caption{Prompt used to merge criteria from multiple survey papers into consolidated criteria. Synthesizes common patterns across different examples. Defined as \texttt{MERGE\_ASPECT\_CRITERIA\_PROMPT}.}
    \label{fig:prompt_merge}
    
\end{figure*}

\begin{figure*}[htbp]
    \centering
    
    \begin{promptbox}{Per-Aspect Outline Evaluation Prompt}
        
        \textbf{System prompt:} You are a domain reviewer for category: \{category\}.
        
        \vspace{0.5em}
        \textbf{Domain-specific outline aspects and their criteria (JSON):}
        
        \{json.dumps(criteria\_payload, ensure\_ascii=False)\}
        
        \vspace{0.5em}
        \textbf{Survey outline (cleaned):}
        
        \{json.dumps(outline, ensure\_ascii=False)\}
        
        \vspace{0.5em}
        \textbf{Instructions:}
        
        For EACH aspect below, give:
        \begin{itemize}
            \item an integer score from 1 to 5
            \item a short justification in one sentence
        \end{itemize}
        
        \vspace{0.5em}
        \textbf{Return JSON exactly in this format:}
        
        \begin{tcolorbox}[
            colback=white,
            colframe=gray!30,
            boxrule=0.5pt,
            sharp corners,
            top=2pt, bottom=2pt
        ]
        \texttt{\{"aspects": [} \\
        \texttt{  \{"aspect\_name": "<string>", "score": <int>, "notes": "<string>"\}} \\
        \texttt{]\}}
        \end{tcolorbox}
        
    \end{promptbox}
    \vspace{-10pt}
    \caption{Prompt for per-aspect evaluation of survey outline quality. Returns individual scores for each aspect, which are then averaged. Used in \texttt{\_build\_prompt\_per\_aspect} method for outline.}
    \label{fig:prompt_eval_outline_peraspect}
    
\end{figure*}

\begin{figure*}[htbp]
    \centering
    
    \begin{promptbox}{Per-Aspect Content Evaluation Prompt}
        
        \textbf{System prompt:} You are a domain reviewer for category: \{category\}.
        
        \vspace{0.5em}
        \textbf{Domain-specific content aspects and their criteria (JSON):}
        
        \{json.dumps(criteria\_payload, ensure\_ascii=False)\}
        
        \vspace{0.5em}
        \textbf{Survey content sections (trimmed for length):}
        
        \{json.dumps(self.\_trim\_sections(content), ensure\_ascii=False)\}
        
        \vspace{0.5em}
        \textbf{Quick stats:}
        
        \{json.dumps(stats, ensure\_ascii=False)\}
        
        \vspace{0.5em}
        \textbf{Instructions:}
        
        For EACH aspect below, give:
        \begin{itemize}
            \item an integer score from 1 to 5
            \item a short justification in one sentence
        \end{itemize}
        
        \vspace{0.5em}
        \textbf{Return JSON exactly in this format:}
        
        \begin{tcolorbox}[
            colback=white,
            colframe=gray!30,
            boxrule=0.5pt,
            sharp corners,
            top=2pt, bottom=2pt
        ]
        \texttt{\{"aspects": [} \\
        \texttt{  \{"aspect\_name": "<string>", "score": <int>, "notes": "<string>"\}} \\
        \texttt{]\}}
        \end{tcolorbox}
        
    \end{promptbox}
    \vspace{-10pt}
    \caption{Prompt for per-aspect evaluation of survey content quality. Returns individual scores for each aspect, which are then averaged. Used in \texttt{\_build\_prompt\_per\_aspect} method for content.}
    \label{fig:prompt_eval_content_peraspect}
    
\end{figure*}

\begin{figure*}[htbp]
    \centering
    
    \begin{promptbox}{Per-Aspect Reference Evaluation Prompt}
        
        \textbf{System prompt:} You are a domain reviewer for category: \{category\}.
        
        \vspace{0.5em}
        \textbf{Domain-specific reference aspects and their criteria (JSON):}
        
        \{json.dumps(criteria\_payload, ensure\_ascii=False)\}
        
        \vspace{0.5em}
        \textbf{Survey reference titles (cleaned):}
        
        \{json.dumps([r.get('title') for r in references], ensure\_ascii=False)\}
        
        \vspace{0.5em}
        \textbf{Instructions:}
        
        For EACH aspect below, give:
        \begin{itemize}
            \item an integer score from 1 to 5
            \item a short justification in one sentence
        \end{itemize}
        
        \vspace{0.5em}
        \textbf{Return JSON exactly in this format:}
        
        \begin{tcolorbox}[
            colback=white,
            colframe=gray!30,
            boxrule=0.5pt,
            sharp corners,
            top=2pt, bottom=2pt
        ]
        \texttt{\{"aspects": [} \\
        \texttt{  \{"aspect\_name": "<string>", "score": <int>, "notes": "<string>"\}} \\
        \texttt{]\}}
        \end{tcolorbox}
        
    \end{promptbox}
    \vspace{-10pt}
    \caption{Prompt for per-aspect evaluation of survey reference quality. Returns individual scores for each aspect, which are then averaged. Used in \texttt{\_build\_prompt\_per\_aspect} method for reference.}
    \label{fig:prompt_eval_reference_peraspect}
    
\end{figure*}

\begin{figure*}[htbp]
    \centering

    \begin{promptbox}{Deep Research Agent Survey Generation Prompt }

        \vspace{0.5em}
        \textbf{Given:}
        \begin{enumerate}
            \item Topic: \{topic\} (user-provided literature review topic)
            \item Output Style: Formal academic style suitable for a top-tier journal
            \item Output Format: A full systematic literature review with section headings and references
        \end{enumerate}

        \vspace{0.5em}
        \textbf{Task (Step 1: Initial Request):}

        Please write a comprehensive systematic literature review paper on: ``\{topic\}''

        The review should cover:
        \begin{enumerate}
            \item Introduction and background
            \item Key concepts and definitions
            \item Historical development and milestones
            \item Current state-of-the-art methods and techniques
            \item Applications and case studies
            \item Challenges and open problems
            \item Future research directions
            \item Conclusion
        \end{enumerate}

        Please write in a formal academic style suitable for a top-tier journal.
        Include section headings and provide references for all sources used.

        \vspace{0.75em}
        \textbf{Task (Step 2: Full Paper Generation):}

        Great. Now please proceed to write the complete literature review as outlined. The review should be comprehensive, well-structured, and rigorous.

        Please include:
        \begin{enumerate}
            \item A clear title and abstract
            \item Introduction with research motivation and objectives
            \item Comprehensive review of existing literature organized by themes or chronology
            \item Critical analysis of the field's strengths and weaknesses
            \item Identification of research gaps
            \item Conclusion summarizing key findings
            \item References section with all cited sources
        \end{enumerate}

        The review should be detailed enough for publication in a reputable academic journal, typically between 3000--8000 words. Please begin writing now.

        \vspace{0.5em}
        \textbf{Notes:}
        \begin{itemize}
            \item The code sends a multi-turn message sequence:
            \begin{enumerate}
                \item User: Step 1 initial request (above)
                \item Assistant: Model ``clarification'' (captured from Step 1 output; if missing, replaced by a short default acknowledgement)
                \item User: Step 2 full-paper instruction (above)
            \end{enumerate}
        \end{itemize}

    \end{promptbox}
    \vspace{-10pt}
    \caption{Prompt for batch generation of systematic literature reviews. The script first issues an initial scope request, then prompts the model to produce a full paper for each user-provided topic; references returned by the deep-research model are collected and saved.}
    \label{fig:prompt_batch}
\end{figure*}

\end{document}